\documentclass[preprint,12pt,numafflabel]{elsarticle}

\usepackage{geometry}
\usepackage{fancyhdr}
\usepackage{titlesec}
\usepackage{float}
\usepackage{subcaption}

\usepackage{amsmath,amsfonts,amssymb, algorithm, algorithmic}
\usepackage{mathtools}
\usepackage{bm}
\usepackage{mathrsfs}

\usepackage{graphicx}
\usepackage{xcolor}
\usepackage{tikz}
\usetikzlibrary{positioning,arrows.meta}
\usepackage{pgfplots}

\usepackage{tabularx}
\usepackage{multirow}
\usepackage{longtable}
\usepackage{colortbl}
\usepackage{booktabs}

\usepackage{enumitem}

\usepackage{hyperref}

\usepackage{natbib}

\usepackage{microtype}
\usepackage{siunitx}
\usepackage{todonotes}
\usepackage{caption}
\usepackage{subcaption}
\usepackage{lipsum}
\usepackage{xspace}
\usepackage{setspace}

\journal{arXiv}
\pgfplotsset{compat=1.18}
\begin{document}
\onehalfspacing

\begin{frontmatter}





\title{Learning Metamaterial Eigenmodes with Wavelet-Encoded Fourier Neural Operators}


\author[Mech]{Han Zhang}
\author[Caltech]{Alexander Ogren}
\author[Comp]{Cynthia Rudin}
\author[Mech]{\protect\\Johann Guilleminot}
\author[Mech]{L. Catherine Brinson}
\affiliation[Mech]{organization={Department of Mechanical Engineering and Materials Science, Duke University},
            city={Durham},
            state={NC},
            country={USA}}
\affiliation[Caltech]{organization={Department of Mechanical Engineering, California Institute of Technology},
            city={Pasadena},
            state={CA},
            country={USA}}
\affiliation[Comp]{organization={Department of Electrical and Computer Engineering, Duke University},
            city={Durham},
            state={NC},
            country={USA}}
\begin{abstract}
Machine learning surrogates based on neural operators have shown broad applicability in solving forward PDE problems. However, eigenvalue problems, in which an eigenparameter and one of several valid eigenmodes must be simultaneously solved, remain difficult because standard operator learning formulations assume a unique input--output map. This work demonstrates that Fourier Neural Operators (FNOs), combined with wavelet-based encodings of PDE inputs, can learn and predict multiple eigenmodes of the elastic wave equation, corresponding to deformation modes of acoustic waves propagating through arbitrary metamaterial geometries. We provide a mechanistic explanation and experimental evidence for why wavelet encodings are well matched to the dual spatial--spectral structure of the FNO, enabling deterministic mode selection on both continuous-valued and binary-valued geometries within a single model, and for why prediction accuracy varies with geometric discontinuities. For metamaterial design, the resulting surrogate accelerates the simulation stage of the design cycle by three orders of magnitude relative to finite element analysis on a consumer-grade CPU, while preserving high fidelity. These results also carry broader implications for designing input encodings in other multi-mode PDE solvers based on spectral neural operators.

\end{abstract}




\end{frontmatter}



\newpage
\section*{Table of Contents}
\begingroup
  \renewcommand{\contentsname}{}
  \tableofcontents
\endgroup
\newpage
\section{Introduction}
\label{introduction}

Acoustic metamaterials are architected materials that enable compact control of sound and elastic waves, including sound attenuation, vibration isolation, and wave focusing, capabilities that are difficult to achieve with homogeneous materials alone~\cite{cummer2016controlling,liao2021acoustic}. These properties arise from geometric structure in addition to material composition, so the dispersive response of a periodic unit cell is governed by the elastic wave equation and is compactly summarized by phononic band structures obtained from Bloch eigenvalue analysis~\cite{kushwaha1993acoustic,hussein2014dynamics}.

Understanding and designing such systems requires computing multiple deformation modes associated with different resonant frequencies and wave behaviors. Traditionally, these modes are obtained through computationally expensive eigenvalue analysis using finite element methods~\cite{Zhang2024uq}. While FEA is highly reliable, this repeated evaluation can make the simulation stage of design cycles prohibitive for optimization, uncertainty quantification, inverse design, and real-time control~\cite{Zhang2024uq,hussein2014dynamics}.

Recent advances in machine learning have introduced neural operators as a promising alternative to traditional numerical solvers. Unlike conventional neural networks that learn mappings between finite-dimensional vectors, neural operators learn mappings between function spaces, allowing them to approximate entire solution operators for families of PDEs~\cite{kovachki2023neural,lu2021deeponet}. Models such as the Fourier Neural Operator (FNO) have demonstrated remarkable accuracy and computational efficiency across a variety of linear and nonlinear PDE problems while maintaining strong generalization capabilities across discretizations and geometries~\cite{li2021fno}. Related physics-informed operator architectures likewise seek rapid emulation of parametric PDE solution maps~\cite{wang2021pidon}. These developments have generated significant interest in using neural operators as surrogate models for computational physics applications.

Despite this progress, existing neural operator architectures have primarily been developed for PDEs in which all coefficients and boundary data are specified and each input corresponds to a unique solution. Eigenvalue PDEs instead require simultaneous recovery of an unknown eigenparameter and one of several valid eigenmodes. In phononic and acoustic metamaterials, prior surrogates have predicted dispersion from material parameters~\cite{liu2019dispersion}, addressed related eigenvalue problems with convolutional networks~\cite{finol2019eigenvalue}, accelerated dispersion evaluation with Gaussian processes~\cite{Ogren2024gpr}, and used neural operators for transmission-loss spectra~\cite{wagner2024neural} or wavevector-conditioned band structures~\cite{liu2026hybridfno}. Deep learning has also been applied to phononic-crystal and metamaterial design~\cite{liu2023deeplearning,jin2022intelligent,muhammad2022ml}, including interpretable methods that extract unit-cell design rules for targeted bandgaps~\cite{CHEN2022101895,Bastawrous2025}. Predicting full Bloch displacement eigenfields while allowing for deterministic multi-mode selection with an FNO remains an unsolved challenge with immense practical benefit.

In this work, we introduce a framework that enables Fourier Neural Operators to solve eigenvalue PDEs associated with elastic wave propagation in acoustic metamaterials. Model inputs are augmented with wavelet-based encodings of PDE parameters that uniquely identify individual eigenmodes while remaining compatible with the FNO architecture. We demonstrate that these encodings allow a single neural operator to learn multiple valid solutions corresponding to distinct deformation modes and to reproduce those modes deterministically on demand. We further show that conventional spatial encodings and purely spectral encodings do not provide the same capability, highlighting the importance of the proposed wavelet representation. We also evaluate the same model on continuous-valued and binary-valued geometries, recovering both displacement fields and eigenfrequencies, and examine how geometric discontinuities interact with the truncated spectral structure of the FNO.

These contributions are summarized as follows:

\begin{enumerate}
    \item We demonstrate that a single Fourier Neural Operator can learn multiple eigenmodes across continuous-valued and binary-valued metamaterial geometries, recovering both eigenvectors (displacement fields) and eigenvalues (eigenfrequencies).
    
    \item We introduce wavelet encodings of modal parameters and explain mechanistically why their spatial--spectral structure enables deterministic mode selection in FNOs.
    
    \item We show that prediction accuracy degrades with geometric discontinuities and interpret this through spectral truncation, with implications for applying FNOs to problems with sharp interfaces.
\end{enumerate}

Together, these results extend neural operators beyond conventional forward PDE problems and provide a pathway to greatly accelerate acoustic metamaterial simulation and design relative to repeated finite element eigenvalue analysis.

\section{Methodology}
\label{sec:Methodology}
This section describes the methods used to generate data and train Fourier Neural Operators for acoustic metamaterial eigenmode prediction. We first define the metamaterial design space and the Bloch FEA simulations used for supervised labels. We then summarize the FNO architecture and the wavelet encodings of modal parameters. Finally, we describe dataset construction and the training procedure.


\subsection{Metamaterial Design Space}
\label{ssec:metamaterial_design_space}
To train a surrogate model capable of predicting acoustic eigenmodes, a representative design space of acoustic metamaterials must first be defined. The design space was selected to provide sufficient geometric and material complexity to demonstrate operator learning for eigenvalue PDEs while maintaining computationally tractable finite element simulations for dataset generation.

The dataset is restricted to two-dimensional acoustic metamaterials composed of infinitely repeating square unit cells exhibiting eight-fold symmetry. Each unit cell is discretized onto a $32\times32$ spatial grid and represented by three material-property channels corresponding to the normalized elastic modulus, density, and Poisson's ratio. Material properties are normalized to the interval $[0,1]$ to improve numerical conditioning during training, while the corresponding physical values are recovered using the fixed material constants listed in Table \ref{tab:parameter_ranges}. The two constituent materials were chosen to represent a structural steel and a stiff elastomer.

\begin{table}[h]
  \centering
  \begin{tabular}{|c|c|c|c|c|c|c|}
    \hline
    $\text{Parameter}$ & $E_{\text{elastomer}}$ & $E_{\text{steel}}$ & $\rho_{\text{elastomer}}$ & $\rho_{\text{steel}}$ & $\nu_{\text{elastomer}}$ & $\nu_{\text{steel}}$ \\
    \hline
    $\text{Units}$ & $\text{Pa}$ & $\text{Pa}$ & $\text{kg/m}^3$ & $\text{kg/m}^3$ & -- & -- \\
    \hline
    $\text{Value}$ & $1 \times 10^{8}$ & $2 \times 10^{11}$ & $1200$ & $8000$ & $0.45$ & $0.3$ \\
    \hline
  \end{tabular}
  \caption{Material parameters. The two constituents are a structural steel and a stiff elastomer.}
  \label{tab:parameter_ranges}
\end{table}

Unit-cell geometries are synthesized by sampling a Gaussian process with a separable periodic covariance on the unit square~\cite{rasmussen2006gaussian}, with period $1$ in each coordinate, signal variance $\sigma_f^2=1$, and length scale $\sigma_l=1$:
\[
k(\mathbf{x},\mathbf{x}')
=
\sigma_f^2
\prod_{d=1}^{2}
\exp\!\left(
-\frac{2\sin^2\!\bigl(\pi|x_d-x_d'|\bigr)}{\sigma_l^2}
\right).
\]
Samples are drawn with mean $0.5$, clipped to $[0,1]$, and then eight-fold ($p4mm$) symmetry is enforced. This procedure produces smoothly varying spatial material distributions compatible with periodically tiled square lattices. The sampled field assumes values between $0$ and $1$, where $0$ represents pure elastomer, $1$ represents pure steel, and intermediate values denote an effective mixture of the two materials below the spatial resolution of the discretization.

To investigate the influence of geometry representation on model performance, two classes of unit cells were generated. The first consists of the continuous valued Gaussian process realizations described above (henceforth referred to as continuous-valued or continuous geometries). The second is obtained by thresholding the Gaussian process samples to binary values, producing geometries composed solely of pure elastomer and pure steel (henceforth referred to as binary-valued or binary geometries). Each representation comprises half of the total dataset.

\begin{figure}[H]
  \centering
  \includegraphics[width=0.5\textwidth]{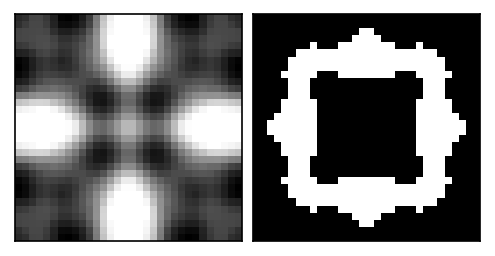}
  \caption{Example of a continuous geometry (left) and a binary geometry (right) generated using the above process and parameters. White pixels represent the stiffer material, black pixels represent the softer material, and grayscale pixels indicate a homogeneous mix of the two.}
  \label{fig:example_geometries}
\end{figure}



\subsection{Wave Propagation Simulations} 
\label{ssec:wave_propagration_simulations}

\begin{figure}[H]
  \centering
  \includegraphics[width=\textwidth]{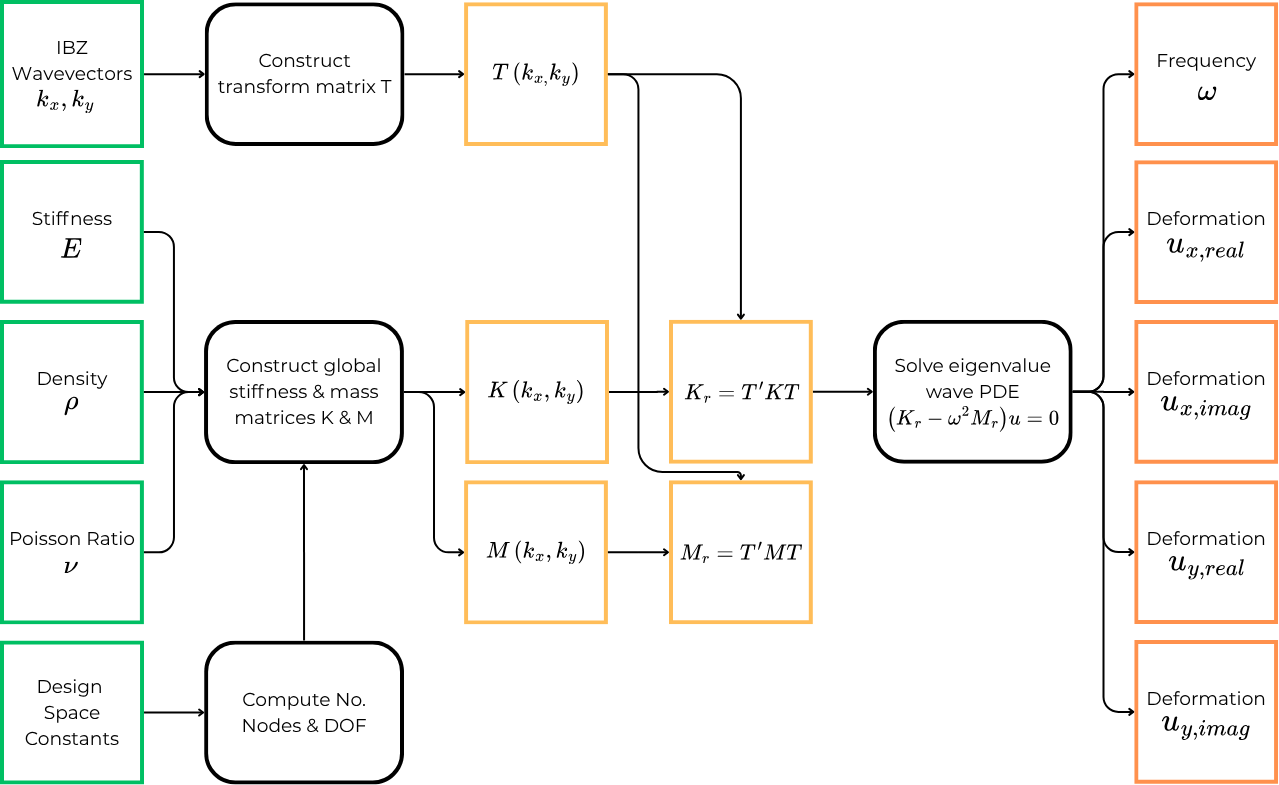}
  \caption{Flowchart of the FEA data generation method. Green squares represent varying inputs to the process, and orange squares represent computed quantities of the FEA process. Black rounded rectangles represent subroutines of the FEA process. The process starts with inputs of defined densities, stiffnesses, and Poisson ratios for each pixel of a metamaterial unit cell geometry (each a $32 \times 32$ pixel grid of values ranging from 0 to 1), along with a chosen $k_x$ and $k_y$ wavevector representing the stimulus to the metamaterial. A custom optimized FEA solver takes these inputs and produces intermediate matrices $(\mathbf{K},\mathbf{M},\mathbf{T})$, forms the reduced operators $(\mathbf{K}_r,\mathbf{M}_r)$, and then solves Eq.~\eqref{eqn:eig_PDE} for the displacement fields (eigenvectors) and frequencies $\omega$ (eigenvalues) for the given inputs.}
  \label{fig:FEA_full_process}
\end{figure}

For each metamaterial geometry, wave propagation is simulated by solving the frequency-domain linear elastodynamic equation with finite element analysis (FEA), following the standard Bloch--Floquet treatment of periodic phononic media~\cite{hussein2014dynamics,kushwaha1993acoustic}. After spatial discretization and enforcement of Bloch periodicity, the governing problem takes the generalized eigenvalue form
\begin{equation}
\label{eqn:eig_PDE}
\left( \mathbf{K}_r(\mathbf{k}) - \omega^2 \mathbf{M}_r(\mathbf{k}) \right) \mathbf{u} = \mathbf{0},
\end{equation}
where $\omega$ is the angular eigenfrequency, $\mathbf{u}$ is the corresponding complex Bloch displacement eigenvector, and $\mathbf{k}$ is the wavevector in the irreducible Brillouin zone (IBZ). The reduced stiffness and mass matrices $\mathbf{K}_r(\mathbf{k})$ and $\mathbf{M}_r(\mathbf{k})$ are obtained from geometry-dependent global finite-element matrices $\mathbf{K}$ and $\mathbf{M}$ via a wavevector-dependent Bloch transformation matrix $\mathbf{T}(\mathbf{k})$,
\begin{equation}
\label{eqn:reduced_KM}
\mathbf{K}_r(\mathbf{k}) = \mathbf{T}(\mathbf{k})^{\dagger}\mathbf{K}\,\mathbf{T}(\mathbf{k}),
\qquad
\mathbf{M}_r(\mathbf{k}) = \mathbf{T}(\mathbf{k})^{\dagger}\mathbf{M}\,\mathbf{T}(\mathbf{k}),
\end{equation}
with $(\cdot)^{\dagger}$ the Hermitian transpose. Here $\mathbf{T}(\mathbf{k})$ maps the full mesh degrees of freedom to an independent set while applying phase factors $\mathrm{e}^{i\mathbf{k}\cdot\mathbf{R}}$ across periodic faces of the unit cell. The matrices $\mathbf{K}$ and $\mathbf{M}$ themselves depend only on the material field and are assembled once per geometry. These intermediate matrices are described briefly so the reader can compare the explicit FEA pipeline (Figure~\ref{fig:FEA_full_process}) with the Fourier neural operator architecture in Section~\ref{ssec:fourier_neural_operator} below (they are not inputs to the neural operator).

The IBZ is discretized into $25$ $x$-direction and $13$ $y$-direction wave numbers, totaling $25 \times 13 = 325$ unique wavevectors covering the IBZ half-plane. For each wavevector, the six lowest eigenfrequencies (bands) were computed using a custom FEA MATLAB script (see Data availability for implementation details). 

Each eigenvector $\mathbf{u}$ consists of two complex-valued displacement fields, $(u_x,u_y)$, which are separated into their real and imaginary components to produce four real-valued output channels, $(u_{x,real},u_{x,imag},u_{y,real},u_{y,imag})$. The corresponding eigenfrequency $\omega$ is encoded as a spatially uniform fifth channel, resulting in a target tensor of size $5\times32\times32$. These five-channel targets are associated with the three-channel material representation from the previous subsection that entered the FEA solve, so each simulation defines the map
\begin{equation*}
\mathbb{R}^{3\times32\times32}
\rightarrow
\mathbb{R}^{5\times32\times32}.
\end{equation*}
The three-channel inputs actually ingested by the Fourier neural operator are assembled differently: a geometry channel is combined with wavelet encodings of the wavevector and band index, as described in Sections~\ref{ssec:input_wavelet_encoding} and~\ref{ssec:dataset_construction}. The overall finite element data-generation procedure is summarized in Figure~\ref{fig:FEA_full_process}.

\subsection{Fourier Neural Operator}
\label{ssec:fourier_neural_operator}

The principal theoretical motivation for Fourier Neural Operators is their universal approximation property over function spaces. Let $\mathcal{A}$ and $\mathcal{U}$ denote Banach spaces of input and output functions, respectively, and let
\begin{equation}
\mathcal{G}:\mathcal{A}\rightarrow\mathcal{U}
\end{equation}
be a continuous nonlinear operator. It has been shown that, given sufficient latent width and Fourier modes, an FNO can approximate $\mathcal{G}$ arbitrarily well on compact subsets of $\mathcal{A}$~\cite{kovachki2023neural}. Unlike conventional neural networks, which approximate finite-dimensional functions, FNOs approximate operators between infinite-dimensional function spaces by learning a sequence of global integral operators parameterized in the Fourier domain~\cite{li2021fno}.

This theoretical framework is formulated for input and output functions in infinite-dimensional spaces. In practice, these functions are represented on a finite computational grid. The continuum Fourier transform is therefore replaced by a discrete Fourier transform (DFT), and only a finite number of Fourier modes are retained during training~\cite{li2021fno,qin2024spectralfno}. Consequently, the learned model approximates the continuum solution operator through its discrete representation while preserving the same spectral operator-learning architecture.

For the present work, the operator acts on discretized fields over a $32\times32$ pixel spatial grid. The input field consists of three channels: a metamaterial geometry channel (the material-location field of Section~\ref{ssec:metamaterial_design_space}, not the three FEA property channels) together with wavelet encodings of the stimulatory wavevector and band index, constructed as detailed in Section~\ref{ssec:input_wavelet_encoding}. The output field consists of five channels corresponding to the predicted eigenvalue, or frequency, and eigenvector, or displacement field, of the resulting deformation mode. Thus, the discretized learned operator takes the form
\begin{equation}
\mathcal{G}_{\theta}:
\mathbb{R}^{32\times32\times3}
\rightarrow
\mathbb{R}^{32\times32\times5}.
\end{equation}

The discretized FNO follows the standard lift--transform--project architecture~\cite{li2021fno}:
\begin{enumerate}
\item \textbf{Lifting:} The input field is embedded into a higher-dimensional latent space through a pointwise lifting operator,
\begin{equation}
v_1(x)=\ell(a(x)),
\qquad
\ell:\mathbb{R}^{3}\rightarrow\mathbb{R}^{d_L}.
\end{equation}

\item \textbf{Fourier layers:} A sequence of Fourier layers is applied to the latent representation. Each layer updates the hidden field according to
\begin{equation}
v_{j+1}(x)
=
\sigma\left(
\mathcal{F}^{-1}
\left[
R_j(k)\mathcal{F}[v_j](k)
\right]
+
W_jv_j(x)
+
b_j
\right),
\end{equation}
where $\mathcal{F}$ denotes the discrete Fourier transform, $R_j(k)$ is the learnable spectral kernel, $W_j$ is a learnable pointwise linear transformation, $b_j$ is a learnable bias term, and $\sigma$ is the activation function. GELU was used in this work because it is smooth and differentiable.

\item \textbf{Projection:} The final latent representation is projected back to the desired output dimension through a pointwise projection operator,
\begin{equation}
u(x)=p(v_n(x)),
\qquad
p:\mathbb{R}^{d_L}\rightarrow\mathbb{R}^{5}.
\end{equation}
\end{enumerate}

FNOs are well suited for acoustic metamaterial simulation because the underlying physics is governed by spatially distributed wave equations, where the response at one location depends on the global geometry of the unit cell and the imposed wavevector. The Fourier layers are designed to use spectral convolutions to efficiently capture long-range interactions and periodic spatial structure, while the pointwise transformations preserve local geometric information. Although FNOs are commonly used as surrogate models for fixed PDE solution operators, the same operator-learning framework can be extended to eigenvalue problems by conditioning the input on modal parameters, such as the wavevector and band index, and training the network to output both the associated eigenvalue and eigenvector field. In this formulation, the FNO learns a parameterized eigensolution operator mapping geometry and modal encodings to the frequency and deformation mode of the acoustic unit cell.

\begin{figure}[H]
\centering
\includegraphics[width=\textwidth]{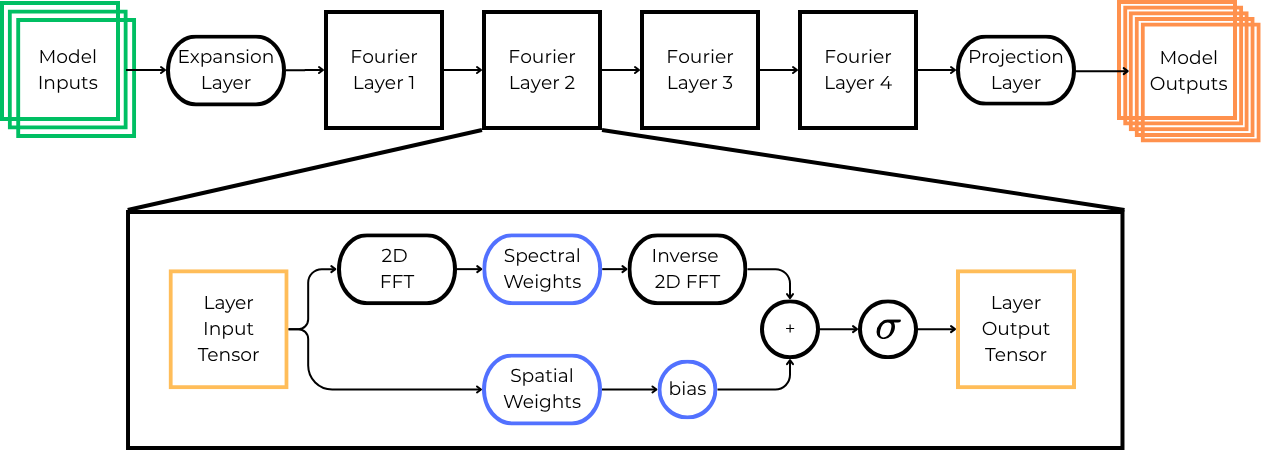}
\caption{The FNO model architecture used in this project. The input data is first expanded according to the hidden-channels parameter, allowing multiple latent features to be represented between Fourier layers. Each Fourier layer is shown in detail in the expanded view. The data passing through a Fourier layer is processed through two learned branches: a pointwise branch, which applies a learned linear transformation in the spatial representation, and a spectral branch, which applies a fast Fourier transform (FFT), multiplies the retained Fourier modes by learned spectral weights, and then applies an inverse FFT. The outputs of both branches, along with a learnable bias term, are summed before passing through a nonlinear activation function. After passing through four Fourier layers, the latent representation is projected from the hidden-channel dimensionality to the final output dimensions.}
\label{fig:FNO_architecture}
\end{figure}

This process is illustrated graphically in Figure~\ref{fig:FNO_architecture}. For our experiments, the NeuralOp Python package was used, which provides FNO model definitions with customizable numbers of layers, hidden channels, and input and output tensor sizes. The hyperparameter combinations explored, as well as the optimized final FNO configuration, are shown in Table~\ref{tab:training_hyperparameters} in Section~\ref{sec:supplement}. These hyperparameters are used alongside the fixed parameters of the problem space: model height $32$, model width $32$, input channels $3$, and output channels $5$.

\subsection{Input Wavelet Encoding}
\label{ssec:input_wavelet_encoding}
To allow the FNO model to toggle between PDE solutions for a given geometry, information must be fed to the model indicating which excitation waveform (wavevectors) and deformation mode (band) we are looking for a solution to. This information is given in the form of wavelet embeddings of the wavevector and band index, which we find to yield lower test losses and better-resolved displacement fields than a constant field embedding (2D matrices of the same value) or a sinusoidal embedding (2D sinusoidal fields with constants encoded as the sinusoidal frequency); quantitative comparisons are reported in Tables~\ref{tab:encoding_ablation_median} and~\ref{tab:encoding_ablation_mean}, and a mechanistic discussion is given in Section~\ref{ssec:band_wavevector_encoding}.

For the wavelet encodings, we use a 1D Gabor wavelet transform to map band numbers to unique wavelet images (see Algorithm \ref{alg:1d_gabor_embedding}), and a 2D Gabor wavelet transform to map x and y wavevectors to another set of unique wavelet images (see Algorithm \ref{alg:2d_gabor_embedding}), inspired by the construction of Gabor wavelets~\cite{gabor1946theory}. 

\begin{figure}[H]
    \centering
    \begin{subfigure}{\textwidth}
        \centering
        \includegraphics[width=\textwidth]{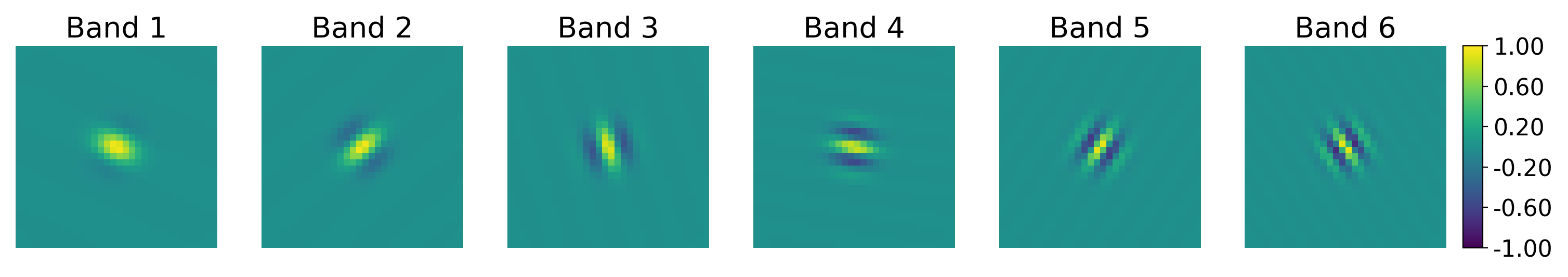}
        \caption{The wavelet spatial encoding for band integer.}
        \label{fig:1D_wavelet_encoding_spatial}
    \end{subfigure}  
    \begin{subfigure}{\textwidth}
        \centering
        \includegraphics[width=\textwidth]{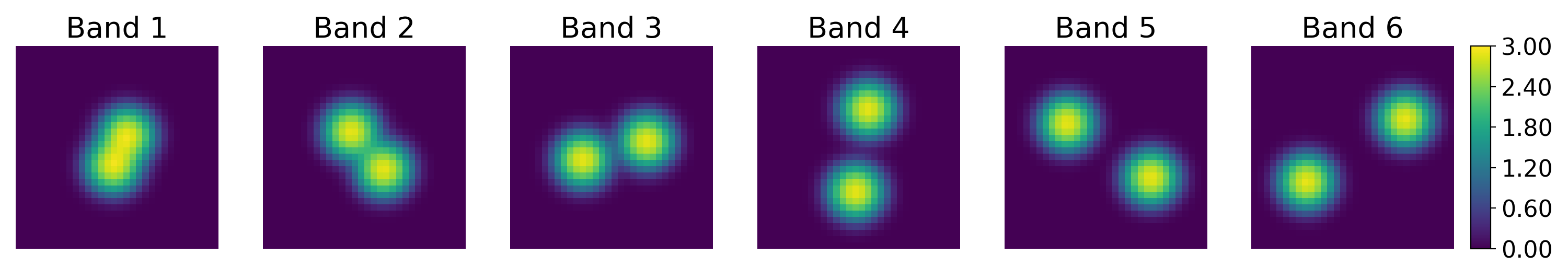}
        \caption{The FFT of the above wavelet spatial encodings.}
        \label{fig:1D_wavelet_encoding_fft}
    \end{subfigure}
    \caption{The wavelet encoding of the band integer (Fig.~\ref{fig:1D_wavelet_encoding_spatial}) and its FFT (Fig.~\ref{fig:1D_wavelet_encoding_fft}). Each band corresponds to a unique wavelet with representation in both spatial and spectral domains.}
    \label{fig:1D_wavelet_encoding}
\end{figure}

\begin{figure}[H]
    \centering
    \begin{subfigure}{\textwidth}
        \centering
        \includegraphics[width=\textwidth]{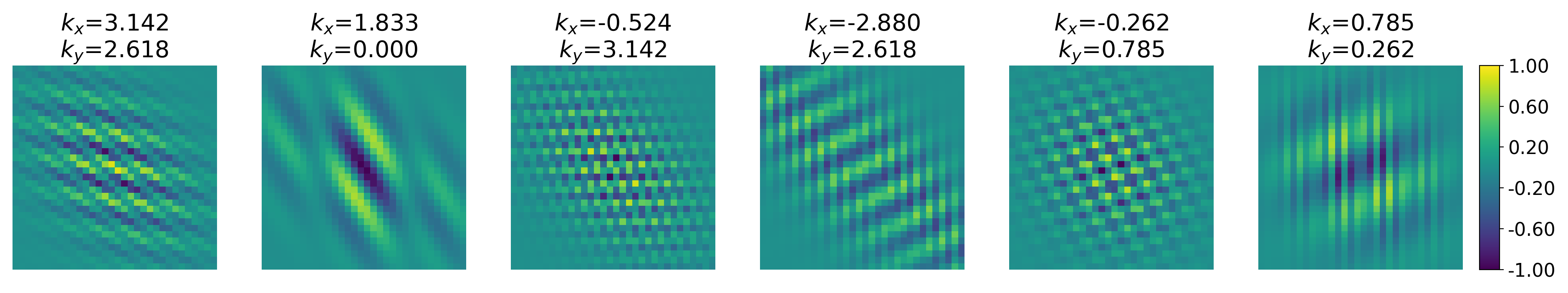}
        \caption{The wavelet spatial encoding for $x$ and $y$ wavevectors.}
        \label{fig:2D_wavelet_encoding_spatial}
    \end{subfigure}  
    \begin{subfigure}{\textwidth}
        \centering
        \includegraphics[width=\textwidth]{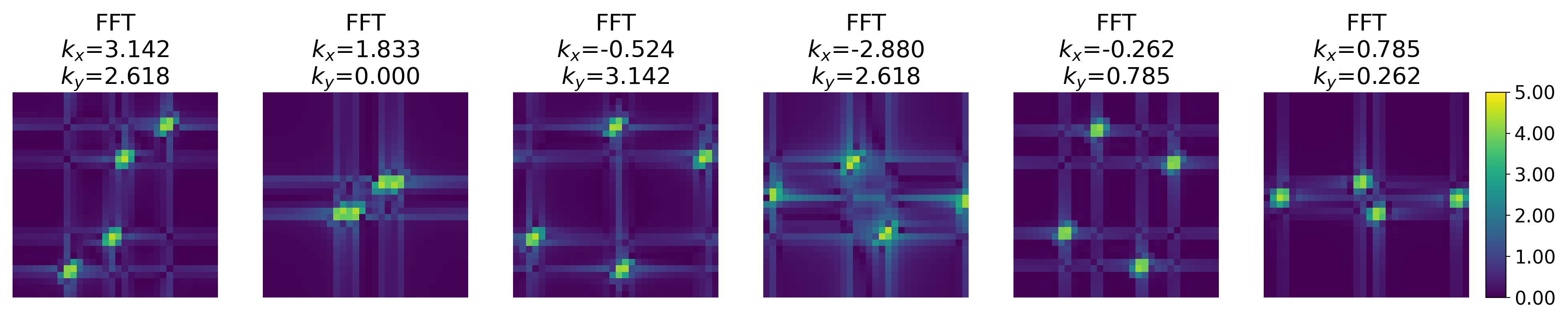}
        \caption{The FFT of the above wavelet spatial encodings.}
        \label{fig:2D_wavelet_encoding_fft}
    \end{subfigure}
    \caption{The wavelet encoding of the $x$ and $y$ wavevectors (Fig.~\ref{fig:2D_wavelet_encoding_spatial}) and its FFT (Fig.~\ref{fig:2D_wavelet_encoding_fft}). The physical wavevectors are shown in the subplot titles. For each $(k_x,k_y)$ pair there is a unique mapping in both the spatial and spectral domains.}
    \label{fig:2D_wavelet_encoding}
\end{figure}

\begin{figure}[H]
  \centering
  \includegraphics[width=0.75\textwidth]{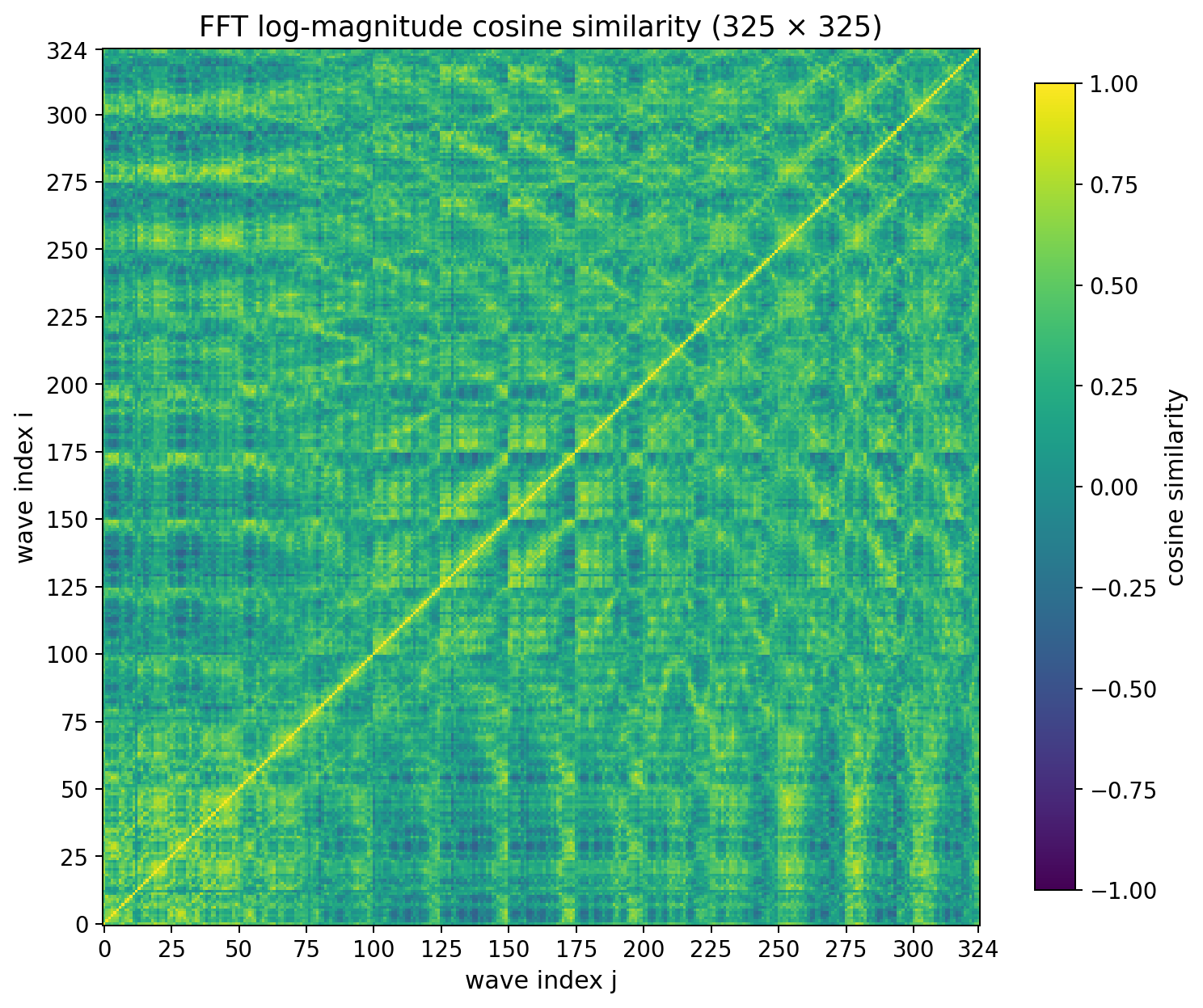}
  \caption{Pairwise cosine similarity $S(\mathbf{k},\mathbf{k}')$ of mean-centered entrywise $\log_{10}|\widehat{\psi}_{\mathbf{k}}|$ spectra for the $325$ IBZ wavevector encodings generated by Algorithm~\ref{alg:2d_gabor_embedding}. The maximum off-diagonal similarity is $0.862$, indicating that distinct $(k_x,k_y)$ pairs remain spectrally separable on the $32\times32$ grid.}
  \label{fig:wavevector_encoding_similarity}
\end{figure}

When encoding continuous constants with oscillatory functions on a finite grid, it is important to verify that distinct parameter values do not alias into nearly identical patches. Otherwise the model cannot distinguish neighboring wavevectors, and mode selection collapses. Let $\psi_{\mathbf{k}}\in\mathbb{R}^{S\times S}$ be the 2D wavelet encoding of wavevector $\mathbf{k}$ as described above, and let $\widehat{\psi}_{\mathbf{k}}\in\mathbb{C}^{S^2}$ be the flattened discrete Fourier transform of that 2D field (FFT computed on the $S\times S$ array, then reshaped to a vector). We define the similarity between two encodings as the cosine similarity of their mean-centered spectral descriptors, which are obtained as follows:
\begin{equation*}
\mathbf{s}_{\mathbf{k}}
=
\log_{10}\lvert\widehat{\psi}_{\mathbf{k}}\rvert
-
\mathrm{mean}\!\left(
\log_{10}\lvert\widehat{\psi}_{\mathbf{k}}\rvert
\right),\quad
S(\mathbf{k},\mathbf{k}')
=
\frac{
\mathbf{s}_{\mathbf{k}}\cdot\mathbf{s}_{\mathbf{k}'}
}{
\|\mathbf{s}_{\mathbf{k}}\|\,\|\mathbf{s}_{\mathbf{k}'}\|
}
\end{equation*}
Evaluating $S$ over all $325$ IBZ wavevectors yields the matrix in Figure~\ref{fig:wavevector_encoding_similarity}. The maximum off-diagonal entry is $0.862$, well below near-duplicate levels, confirming that the chosen encoding constants introduce no appreciable aliasing across the IBZ grid.

\subsection{Dataset Construction} 
\label{ssec:dataset_construction}
Our full dataset contains 24{,}000 geometries, randomly generated as described in prior subsections. Half are binary geometries, with only 0s or 1s in each pixel location representing one of the two materials, while the other half are continuous geometries, with any value in $[0,1]$ at each pixel representing a blended material. Both geometry classes are combined into one training pool and learned simultaneously by a single model. Each geometry has six bands and 325 wavevectors per band (a mesh grid of the IBZ half-plane), resulting in $24000 \times 6 \times 325 = 46{,}800{,}000$ sample pairs of inputs with shape $(3 \times 32 \times 32)$ and outputs of shape $(5 \times 32 \times 32)$. The data are stored as float16 PyTorch tensors. 


\begin{figure}[H]
  \centering
  \begin{subfigure}[c]{0.5\textwidth}
    \centering
    \includegraphics[scale=0.35]{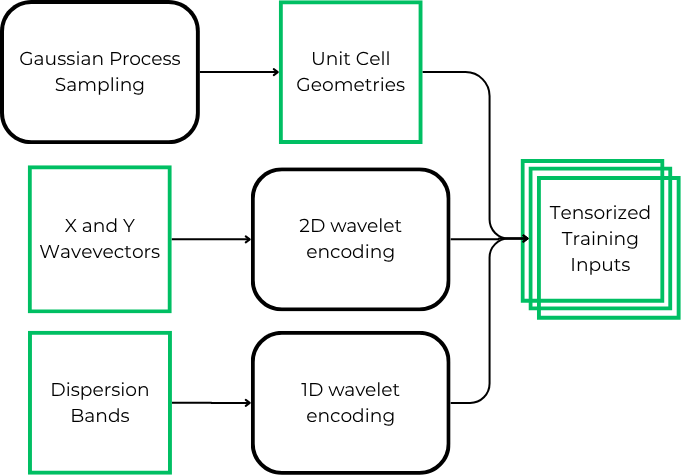}
    \caption{Construction of the input tensors}
    \label{fig:input_dataset_construction}
  \end{subfigure}%
  \hfill
  \begin{subfigure}[c]{0.5\textwidth}
    \centering
    \includegraphics[scale=0.35]{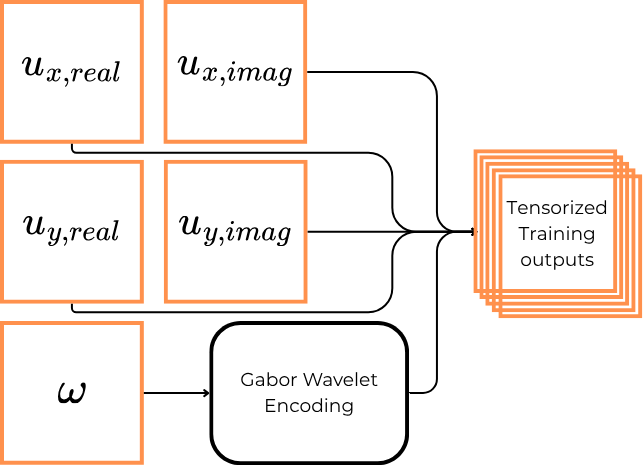}
    \caption{Construction of the output tensors}
    \label{fig:output_dataset_construction}
  \end{subfigure}
  \caption{Subfigures~\subref{fig:input_dataset_construction} and~\subref{fig:output_dataset_construction} show the processes by which input data from Section~\ref{ssec:metamaterial_design_space} and output data from Section~\ref{ssec:wave_propagration_simulations} are tensorized for ingestion by the FNO model. Note the wavelet embedding steps applied to fundamentally scalar quantities.}
  \label{fig:dataset_construction}
\end{figure}



\subsection{Training Procedure}
\label{ssec:training_procedure}
The Fourier Neural Operator was trained using supervised learning on paired inputs and outputs described in Sections~\ref{ssec:dataset_construction} and~\ref{ssec:input_wavelet_encoding}. Each training sample consists of a $3 \times 32 \times 32$ input tensor (geometry and wavelet embeddings) and a $5 \times 32 \times 32$ output tensor (displacement field components and eigenfrequency). A supplementary encode--decode fidelity check for positive scalars---distinct from the input band and wavevector encodings---is reported in Section~\ref{ssec:wavelet_decoding}.

Five loss functions were evaluated as training objectives: mean absolute error (MAE), mean squared error (MSE), normalized MAE (NMAE), normalized MSE (NMSE), and structural similarity index (SSIM). Losses were aggregated across all five output channels. NMAE yielded the best aggregate performance and was used for all reported results. Its definition is given below and details on the other loss functions are provided in the Supplementary Information.
\begin{equation*}
    \mathcal{L}_{\text{NMAE}} = \frac{1}{HW} \sum_{i=0}^{H-1} \sum_{j=0}^{W-1} \sum_{c=0}^{4} \frac{\displaystyle \left| \hat{u}_{c}(i,j) - u_{c}(i,j) \right|}{\left| u_{c}(i,j) \right| + \varepsilon}
\end{equation*}

where $H = W = 32$ (pixels), $\hat{u}$ denotes the model prediction, $u$ denotes the ground-truth target, $\varepsilon$ is a small stabilizer, and $c \in \{0,1,2,3,4\}$ corresponds to the 5 output channels: eigenfrequency, followed by real and imaginary components of x and y displacements.

For training the FNO model, combinations of model layers, hidden channels, activation function, loss function, optimizer, and training hyperparameters were evaluated. The AdamW optimizer~\cite{loshchilov2019adamw} achieved the best performance compared with Adam~\cite{kingma2015adam}, other momentum-based optimizers, and SGD. The candidate settings and selected values are reported in Table~\ref{tab:training_hyperparameters} in Section~\ref{sec:supplement}, with the selected combination values italicized.

The selected architecture uses four Fourier layers, 128 hidden channels, and GELU activations. Increasing the depth or width beyond these values did not appreciably improve validation performance and increased both training cost and the tendency to overfit. Final model performance did not appreciably change with variations in scheduler step size or batch size trialed, so these parameters were fixed at 1 and 520, respectively. Higher learning rates accelerated convergence but produced less stable optimization. In particular, learning rates above approximately $10^{-2}$ occasionally caused an abrupt, irreversible loss spike, after which training plateaued. The final selected training configuration uses NMAE throughout, AdamW optimization, a learning rate of $2\times10^{-3}$, a batch size of 520, and 12 epochs.

\section{Results and Discussion}

The results and discussion are organized as follows. We first assess Bloch displacement predictions on continuous and binary geometries. Then, we showcase eigenfrequency and dispersion accuracy. From here, we examine how prediction error on binary geometries tracks material interface length. Finally, we compare wavelet, sinusoidal, and uniform encodings effects on model performance under matched training settings.

\subsection{Displacement Field Prediction}
We first evaluate the trained FNO on its predictions of the Bloch displacement fields $(u_x,u_y)$. For each unseen test sample, the prediction error is measured by the normalized mean absolute error (NMAE) over the four real-valued displacement channels, $(u_{x,\mathrm{real}},u_{x,\mathrm{imag}},u_{y,\mathrm{real}},u_{y,\mathrm{imag}})$. Samples are ranked by performance, defined inversely with NMAE, so larger losses correspond to lower performance percentiles. We select representative cases at the $25$th, $50$th, and $75$th performance percentiles, corresponding to relatively weak, median, and relatively strong predictions. The same percentile comparison is shown for both the binary and continuous geometry datasets, allowing direct visual assessment of how well the model recovers the spatial structure of the displacement modes across contrasting geometry regimes.

\subsubsection{Continuous Geometries}
\begin{figure}[H]
    \centering
    \begin{subfigure}[c]{0.75\textwidth}
        \centering
        \includegraphics[width=\textwidth]{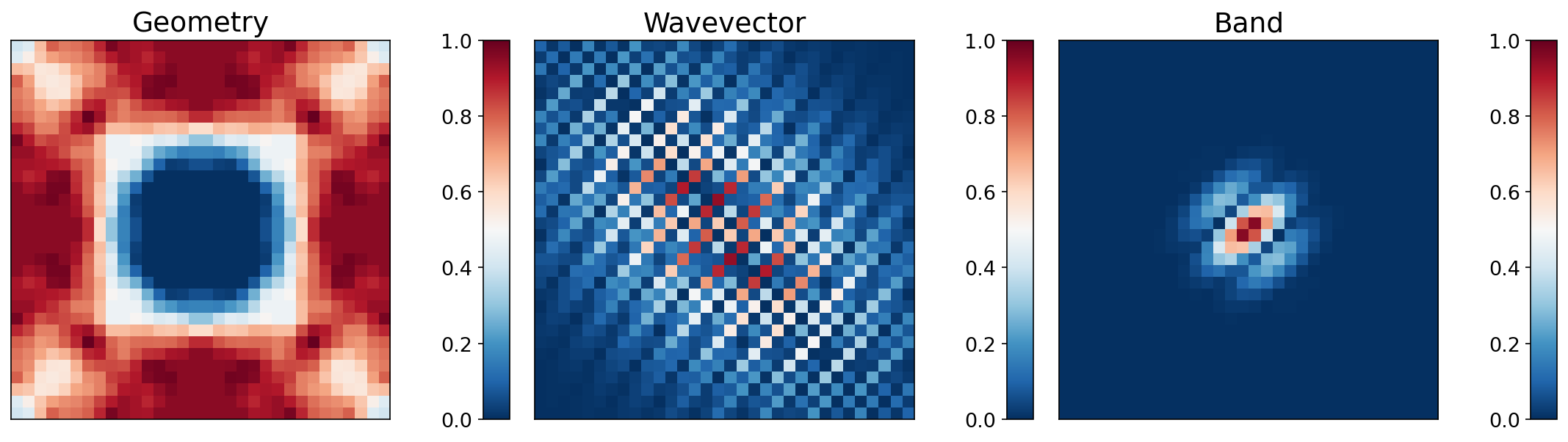}
        \caption{Input sample at the 25th percentile of performance for unseen continuous geometries.}
        \label{fig:c_p25_input}
    \end{subfigure}
    \vspace{1em}
    \begin{subfigure}[c]{\textwidth}
    \centering
        \includegraphics[width=\textwidth]{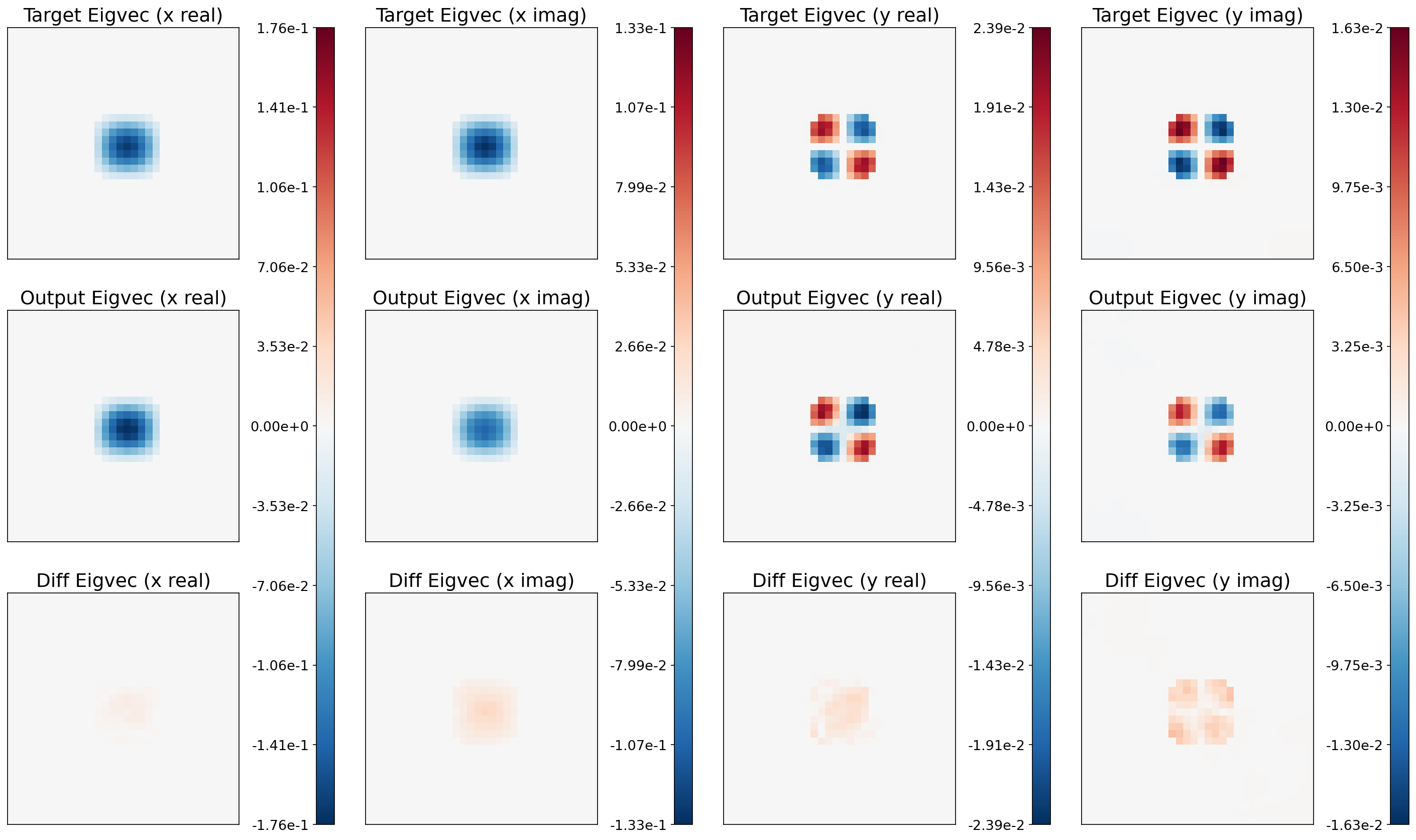}
        \caption{Output predictions at the 25th percentile of performance for unseen continuous geometries. The first row shows target fields, the second row shows predictions, and the third row shows the differences between targets and predictions. At the 25th percentile, prediction outcomes are already fairly good, with accurate representations of scale and structure.}
        \label{fig:c_p25_output}
    \end{subfigure}
\end{figure}

\begin{figure}[H]
    \centering
    \begin{subfigure}[c]{0.75\textwidth}
        \centering
        \includegraphics[width=\textwidth]{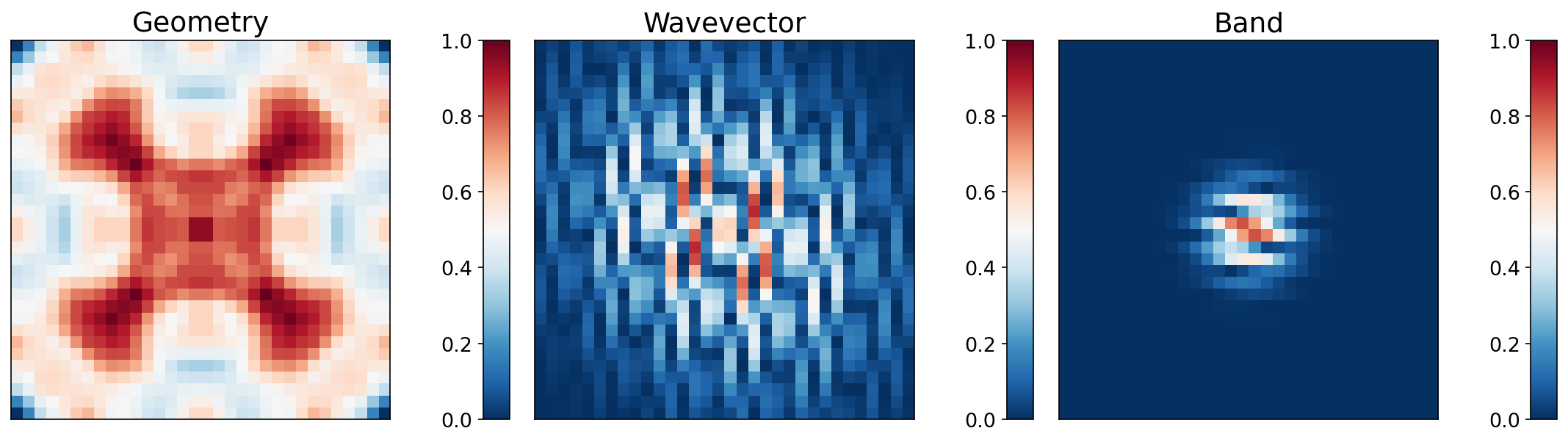}
        \caption{Input sample at the 50th percentile of performance for unseen continuous geometries.}
        \label{fig:c_p50_input}
    \end{subfigure}
    \vspace{1em}
    \begin{subfigure}[c]{\textwidth}
    \centering
        \includegraphics[width=\textwidth]{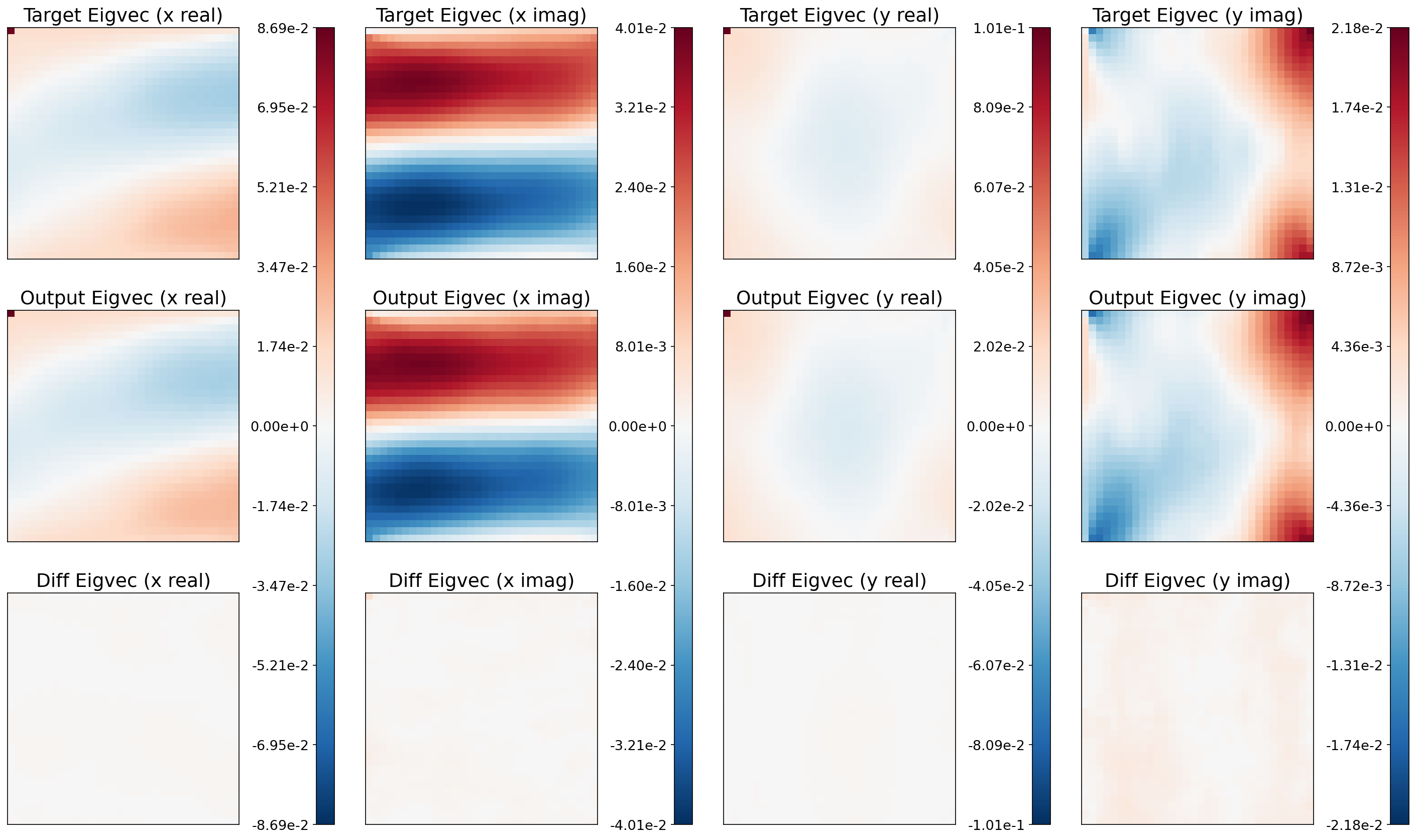}
        \caption{Output predictions at the 50th percentile of performance for unseen continuous geometries. The first row shows target fields, the second row shows predictions, and the third row shows the differences between targets and predictions. Predictions are able to consistently capture complex structures without graininess.}
        \label{fig:c_p50_output}
    \end{subfigure}
\end{figure}

\begin{figure}[H]
    \centering
    \begin{subfigure}[c]{0.75\textwidth}
        \centering
        \includegraphics[width=\textwidth]{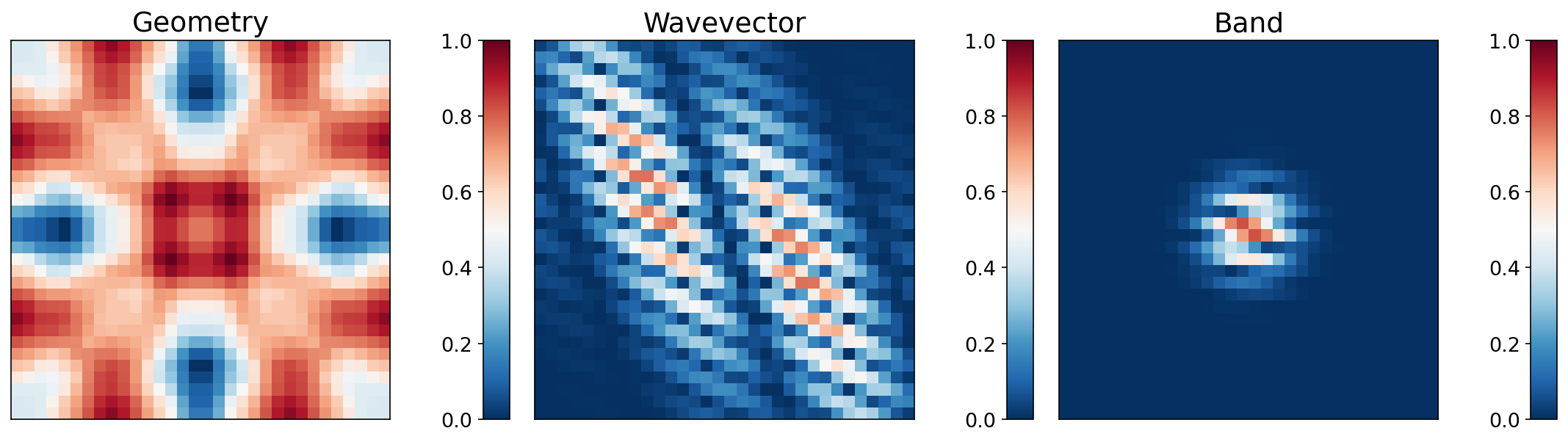}
        \caption{Input sample at the 75th percentile of performance for unseen continuous geometries.}
        \label{fig:c_p75_input}
    \end{subfigure}
    \vspace{1em}
    \begin{subfigure}[c]{\textwidth}
    \centering
        \includegraphics[width=\textwidth]{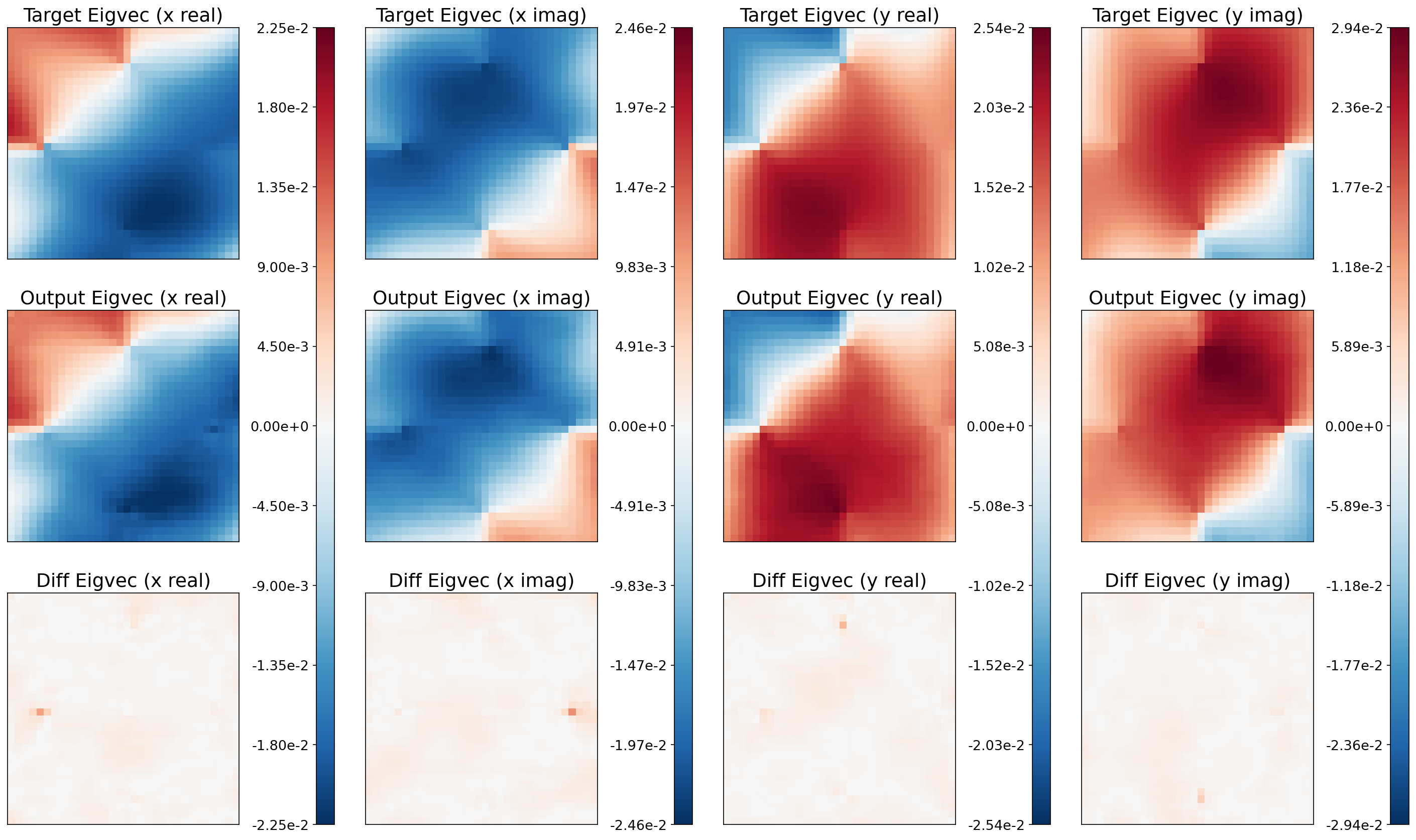}
        \caption{Output predictions at the 75th percentile of performance for unseen continuous geometries. The first row shows target fields, the second row shows predictions, and the third row shows the differences between targets and predictions. At this point, the target and prediction are virtually indistinguishable by eye.}
        \label{fig:c_p75_output}
    \end{subfigure}
\end{figure}
These results demonstrate that with wavelet encodings the FNO performs well on continuous geometries, faithfully reproducing the displacement fields in most cases. Model predictions in the upper performance quartiles match the targets well in structure and scale, with average pixel relative errors decreasing smoothly from on the order of $10^{-2}$ to $10^{-4}$ as performance percentile increases (see Figure~\ref{fig:disp_loss_histogram_continuous} for detailed statistics).

\subsubsection{Binary Geometries}
\begin{figure}[H]
    \centering
    \begin{subfigure}[c]{\textwidth}
        \centering
        \includegraphics[width=0.75\textwidth]{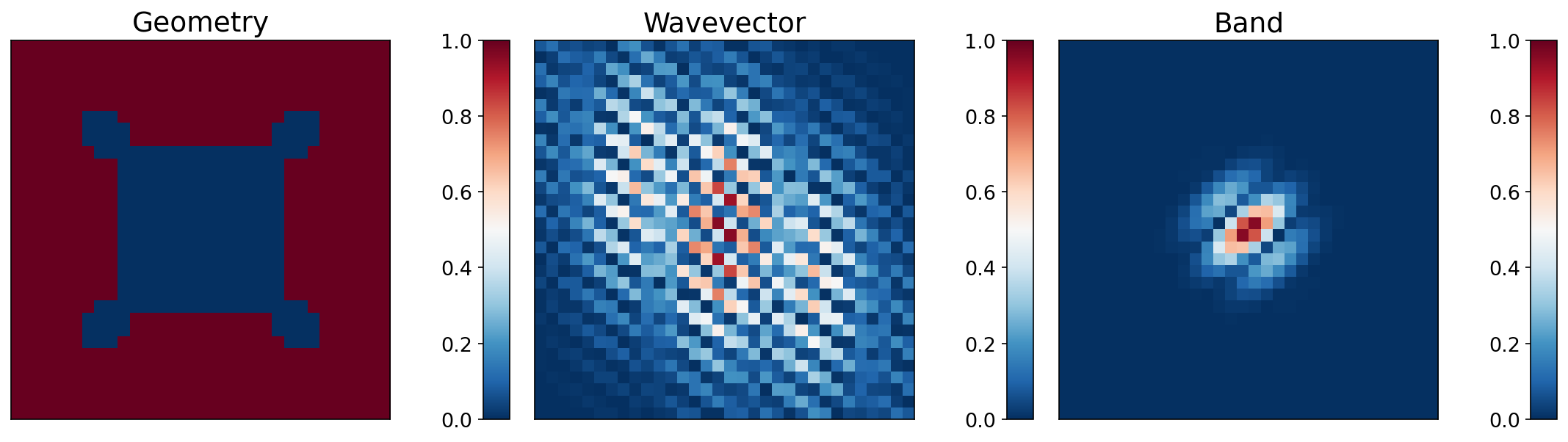}
        \caption{Input sample at the 25th percentile of performance for unseen binary geometries.}
        \label{fig:b_p25_input}
    \end{subfigure}
    \vspace{1em}
    \begin{subfigure}[c]{\textwidth}
    \centering
        \includegraphics[width=\textwidth]{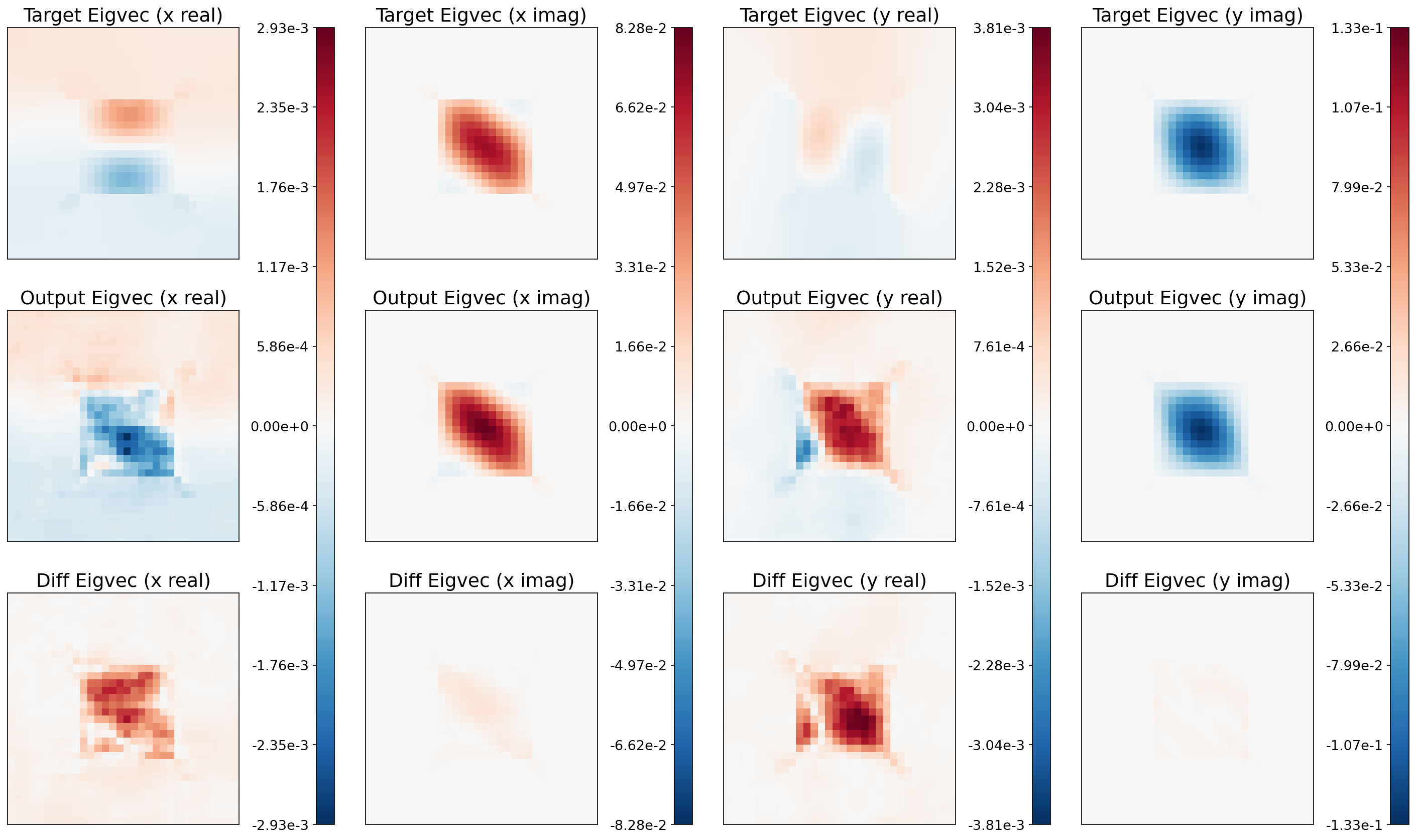}
        \caption{Output predictions at the 25th percentile of performance for unseen binary geometries. The first row shows target fields, the second row shows predictions, and the third row shows the differences between targets and predictions. Because NMAE normalizes residuals by target magnitude, low-amplitude channels receive greater relative weight during training than their absolute scale would suggest. For this sample the imaginary displacement channels are one to two orders of magnitude larger than the real channels, so absolute error can remain low overall while relative error stays large on the smaller-magnitude channels. (Note that the colorbars for each column are independently scaled.)}
        \label{fig:b_p25_output}
    \end{subfigure}
\end{figure}

\begin{figure}[H]
    \centering
    \begin{subfigure}[c]{\textwidth}
        \centering
        \includegraphics[width=0.75\textwidth]{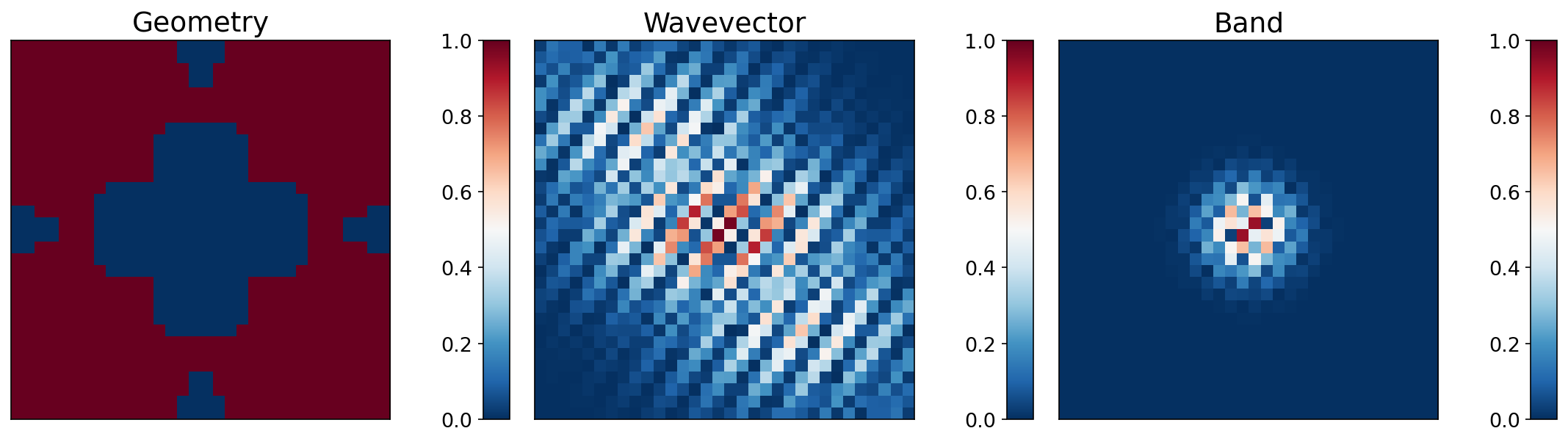}
        \caption{Input sample at the 50th percentile of performance for unseen binary geometries.}
        \label{fig:b_p50_input}
    \end{subfigure}
    \vspace{1em}
    \begin{subfigure}[c]{\textwidth}
    \centering
        \includegraphics[width=\textwidth]{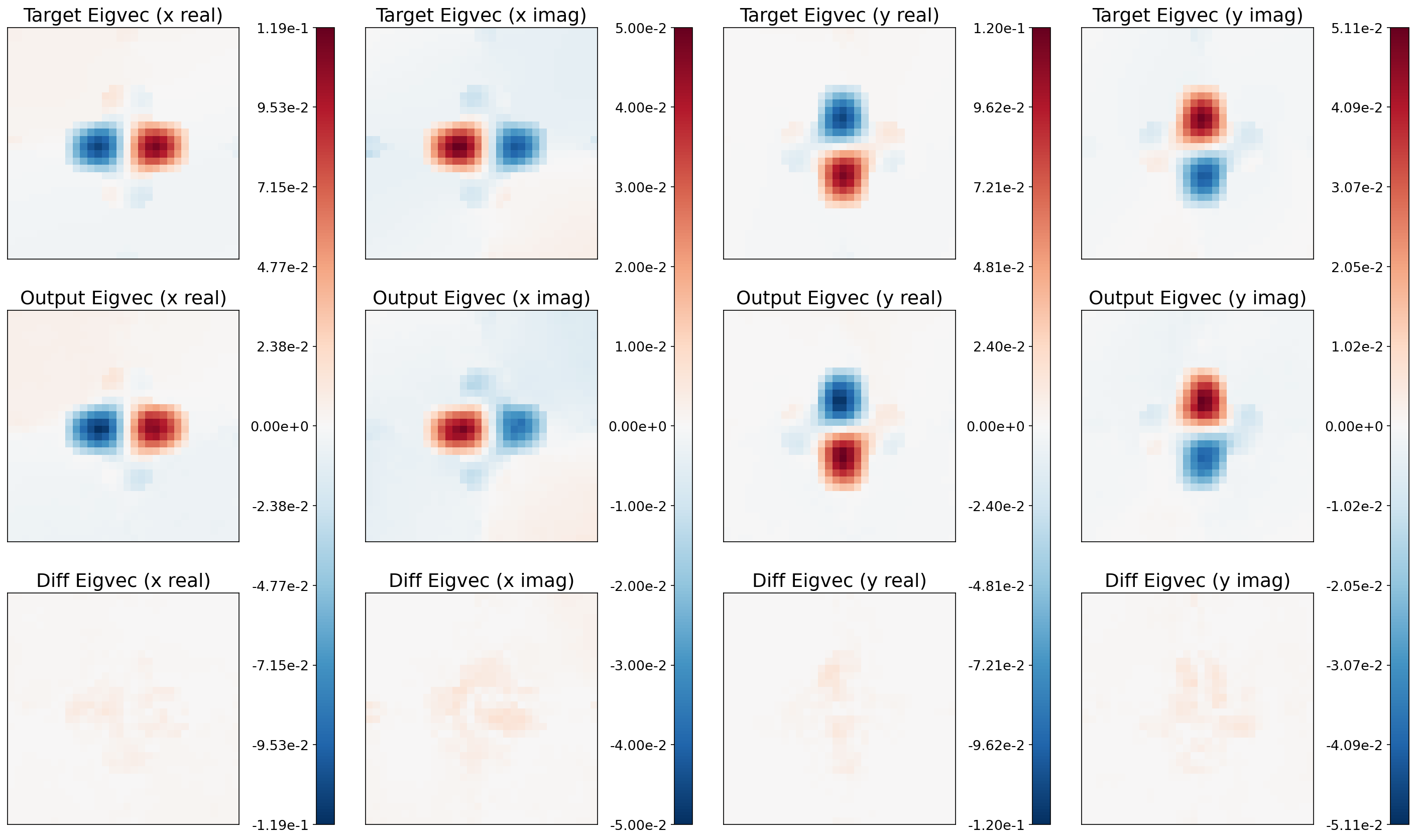}
        \caption{Output predictions at the 50th percentile of performance for unseen binary geometries. The first row shows target fields, the second row shows predictions, and the third row shows the differences between targets and predictions. Absolute and relative errors are low, with strong agreement in structure and scale between targets and predictions.}
        \label{fig:b_p50_output}
    \end{subfigure}
\end{figure}

\begin{figure}[H]
    \centering
    \begin{subfigure}[c]{\textwidth}
        \centering
        \includegraphics[width=0.75\textwidth]{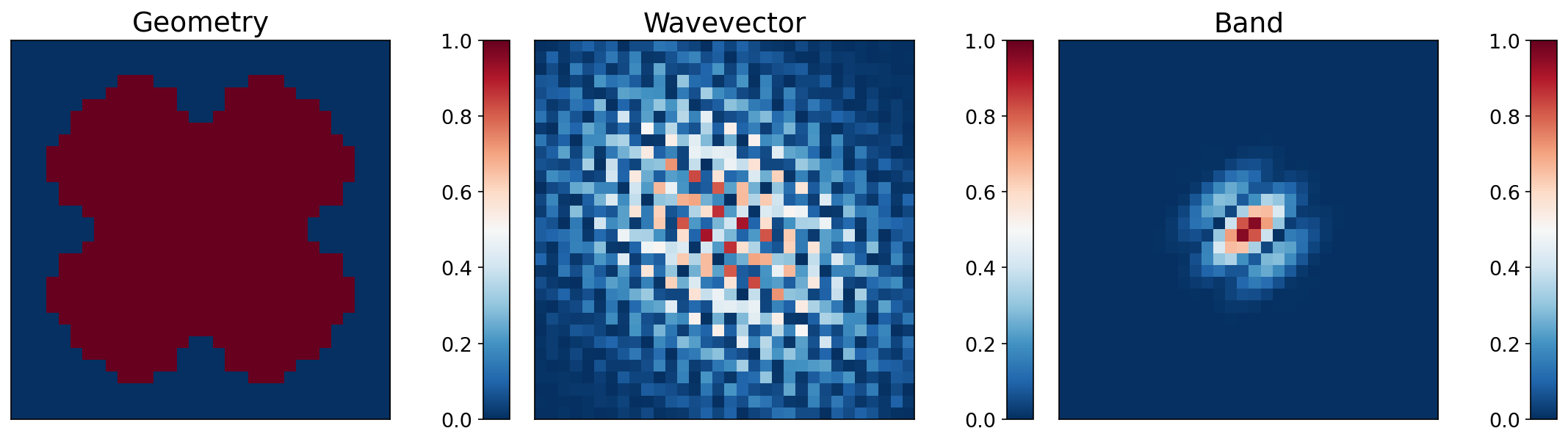}
        \caption{Input sample at the 75th percentile of performance for unseen binary geometries.}
        \label{fig:b_p75_input}
    \end{subfigure}
    \vspace{1em}
    \begin{subfigure}[c]{\textwidth}
    \centering
        \includegraphics[width=\textwidth]{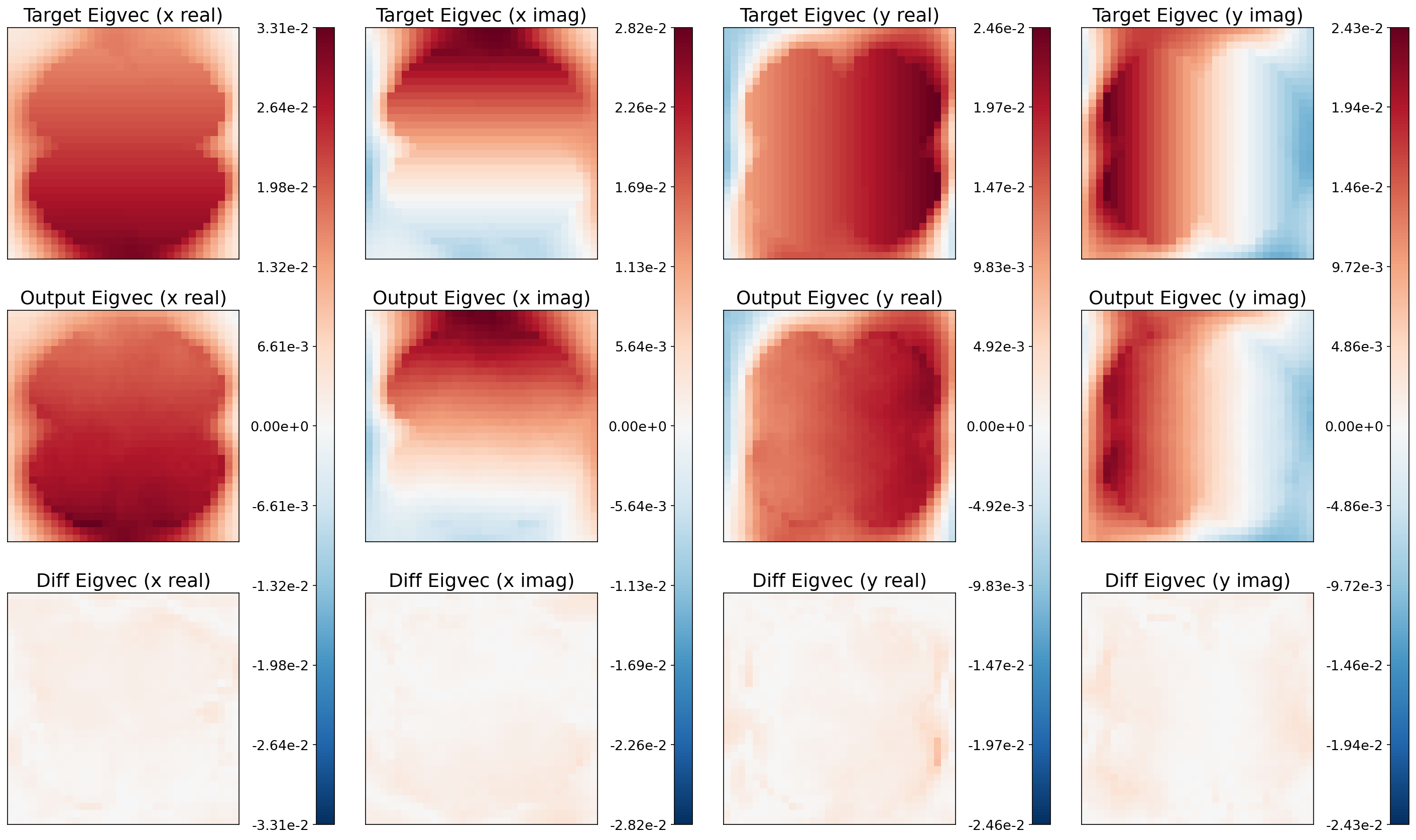}
        \caption{Output predictions at the 75th percentile of performance for unseen binary geometries. The first row shows target fields, the second row shows predictions, and the third row shows the differences between targets and predictions. At this point, disagreements between targets and predictions become very hard to see by eye.}
        \label{fig:b_p75_output}
    \end{subfigure}
\end{figure}

As in the continuous geometry case, these results demonstrate that with wavelet encodings the FNO performs well on binary geometries, faithfully reproducing the displacement fields in most cases. Model predictions in the upper performance quartiles match the targets well in structure and scale, with average pixel relative errors decreasing smoothly from on the order of $10^{-2}$ to $10^{-4}$ as performance percentile increases (see Figure~\ref{fig:disp_loss_histogram_discrete} for detailed statistics).

\subsubsection{Continuous Versus Binary Performance}
\begin{figure}[H]
    \centering
    \begin{subfigure}[t]{0.49\textwidth}
        \centering
        \includegraphics[width=\textwidth]{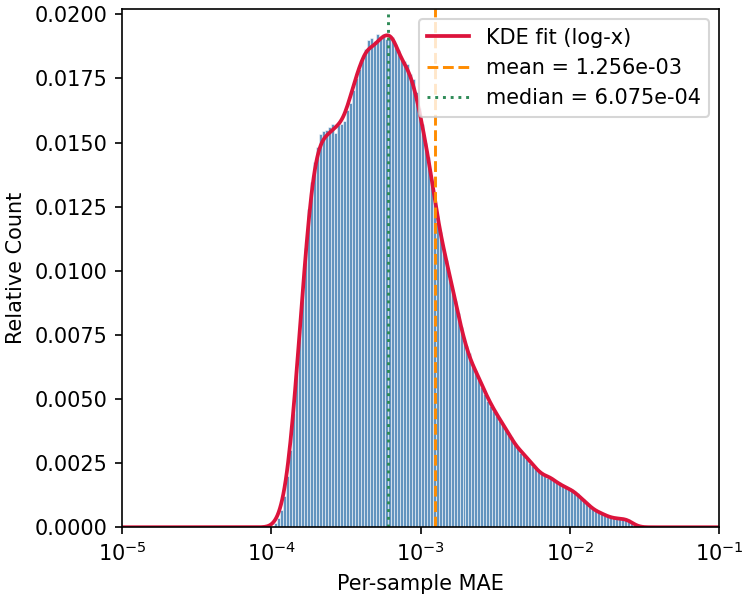}
        \caption{Performance histogram for continuous geometries.}
        \label{fig:disp_loss_histogram_continuous}
    \end{subfigure}
    \hfill
    \begin{subfigure}[t]{0.49\textwidth}
        \centering
        \includegraphics[width=\textwidth]{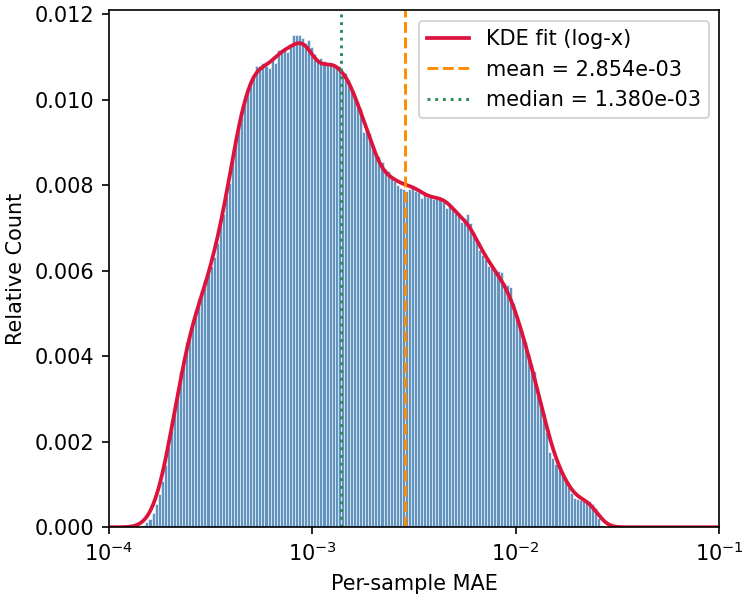}
        \caption{Performance histogram for binary geometries.}
        \label{fig:disp_loss_histogram_discrete}
    \end{subfigure}
    \caption{Comparison of relative prediction error distributions for continuous and binary metamaterial geometries for the same model. While there are some tail outliers, most samples fall within the range of $~10^{-2}$ to $~10^{-4}$, indicating that the same model is able to learn continuous and binary geometry cases.}
    \label{fig:disp_loss_histograms}
\end{figure}

A single FNO with wavelet encodings being able to perform well on both continuous and binary geometries is a notable result. Encoding comparisons that outline theoretical reasons for this performance are given in Section~\ref{ssec:band_wavevector_encoding}. Model performance over all samples in the test set is shown in Figure~\ref{fig:disp_loss_histograms}. 

The poorer performance on binary geometries is consistent with the preference of Fourier-parameterized networks for smoother inputs and outputs~\cite{rahaman2019spectral,qin2024spectralfno}: binary designs introduce jump discontinuities and broadband high-wavenumber content that a truncated spectral representation resolves less faithfully. We develop this spectral-truncation account and its empirical support in Section~\ref{ssec:boundary_length_vs_loss}.

These results indicate two important takeaways. The first is that FNOs with wavelet encodings can learn multiple deformation modes for each geometry, which the user can select by setting the wavevector and band inputs. The second is that a single trained model is robust across continuous and binary geometries from the same design space, rather than requiring separate specialization for each, which suggests that the mapping captures shared structure in the underlying eigenvalue map. At the same time, Section~\ref{ssec:boundary_length_vs_loss} shows that accuracy tends to degrade for more discontinuous binary designs: as material interfaces lengthen and neighboring pixels change more abruptly, spectral truncation on the fixed grid makes the problem harder. Whether model robustness extends to distinct design spaces or symmetry classes is left for future work.

\subsection{Dispersion Band Reconstruction}
\begin{figure}[H]
    \centering
    \begin{subfigure}[t]{0.32\textwidth}
        \centering
        \includegraphics[width=\textwidth]{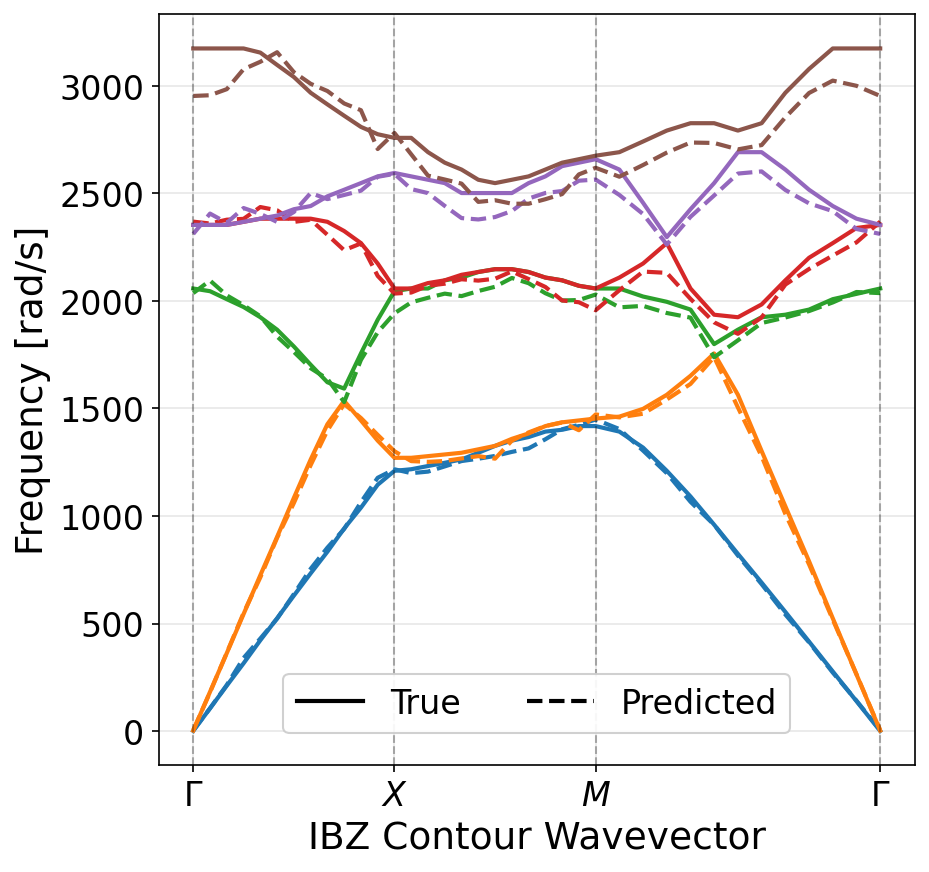}
        \caption{25th percentile performance.}
        \label{fig:dispersion_c_p25}
    \end{subfigure}
    \hfill
    \begin{subfigure}[t]{0.32\textwidth}
        \centering
        \includegraphics[width=\textwidth]{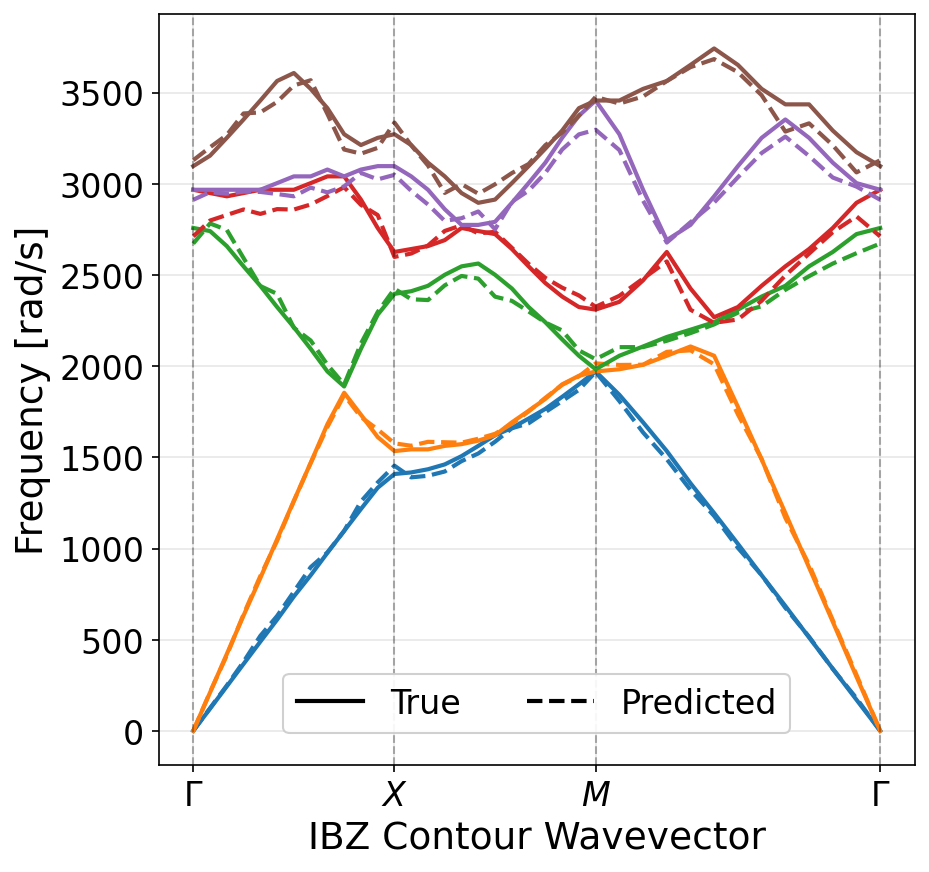}
        \caption{50th percentile performance.}
        \label{fig:dispersion_c_p50}
    \end{subfigure}
    \hfill
    \begin{subfigure}[t]{0.32\textwidth}
        \centering
        \includegraphics[width=\textwidth]{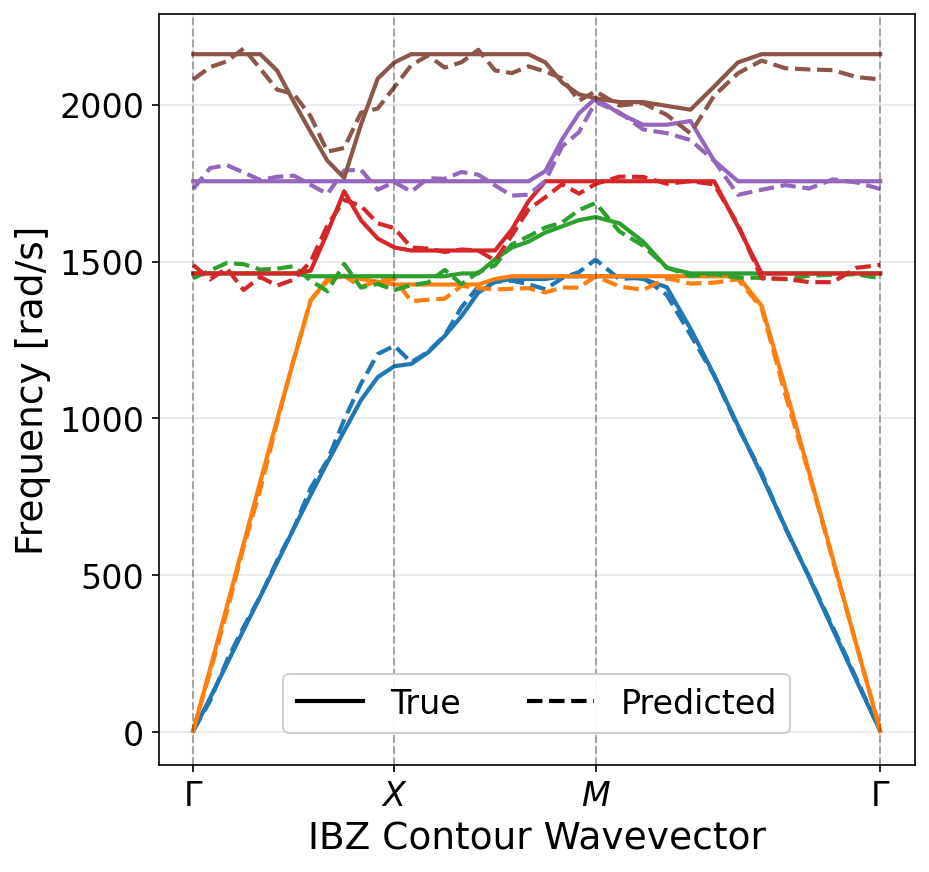}
        \caption{75th percentile performance.}
        \label{fig:dispersion_c_p75}
    \end{subfigure}
    \caption{Predicted and ground-truth dispersion bands for representative unseen continuous geometries spanning the 25th, 50th, and 75th percentiles of prediction performance, restricted to the horizontal, vertical, and diagonal IBZ traversals.}
    \label{fig:dispersion_c_percentiles}
\end{figure}
These plots show eigenfrequency predictions across all 325 wavevectors of the IBZ half-plane for continuous geometries, restricted to the horizontal, vertical, and diagonal IBZ traversals. Because the eigenfrequency is encoded with a logarithmic transform to support multi-scale learning, errors tend to scale with the target magnitude, and higher bands can appear more deviant on a linear plot. Throughout, ``encoded eigenfrequency'' refers to this log-transformed, spatially uniform output channel rather than physical frequency in Hz. Overall, the model achieves an average prediction error below $1\%$ on the encoded eigenfrequency across unseen samples.

\begin{figure}[H]
    \centering
    \begin{subfigure}[t]{0.32\textwidth}
        \centering
        \includegraphics[width=\textwidth]{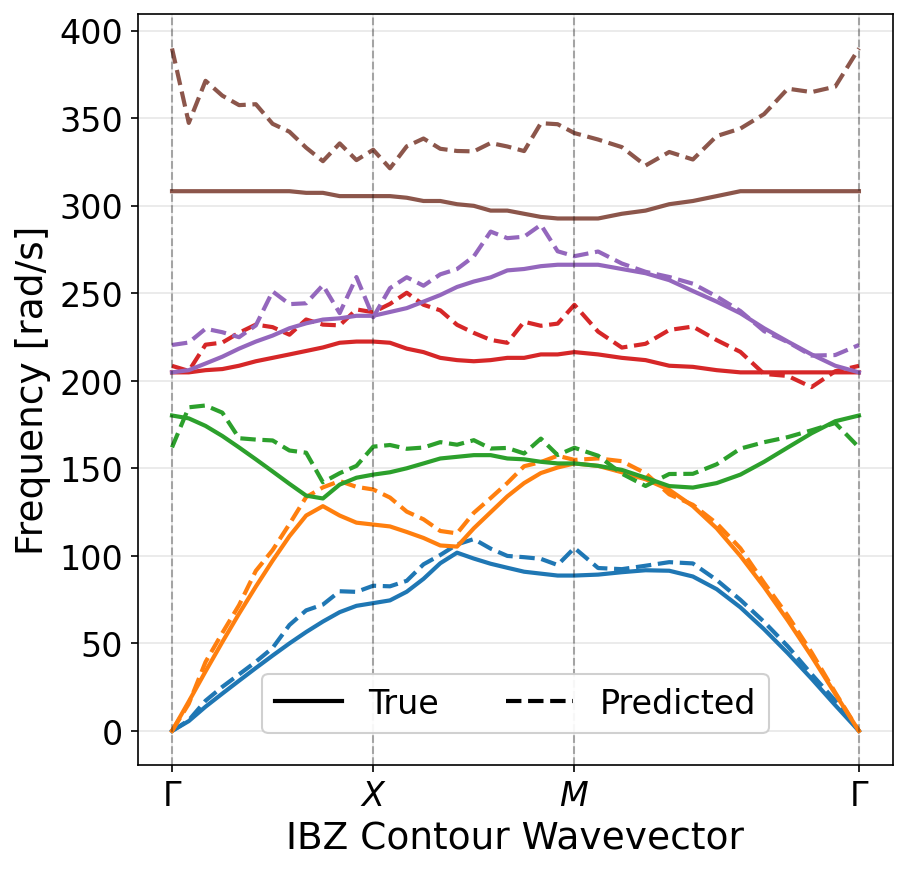}
        \caption{25th percentile performance.}
        \label{fig:dispersion_b_p25}
    \end{subfigure}
    \hfill
    \begin{subfigure}[t]{0.32\textwidth}
        \centering
        \includegraphics[width=\textwidth]{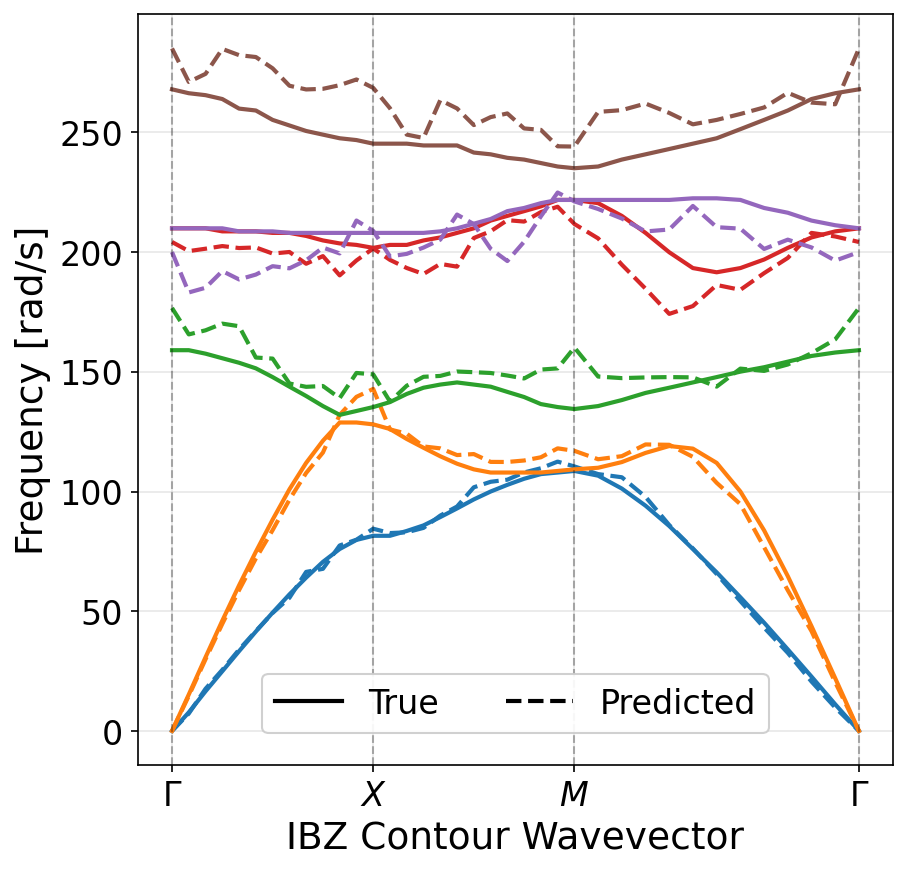}
        \caption{50th percentile performance.}
        \label{fig:dispersion_b_p50}
    \end{subfigure}
    \hfill
    \begin{subfigure}[t]{0.32\textwidth}
        \centering
        \includegraphics[width=\textwidth]{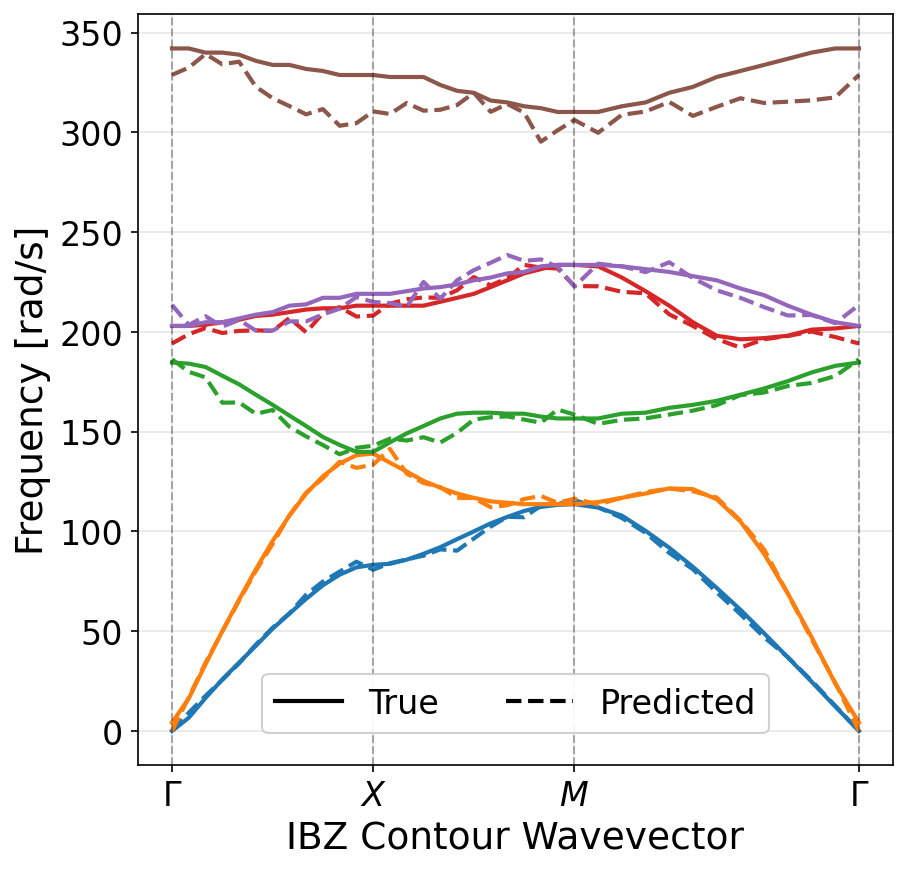}
        \caption{75th percentile performance.}
        \label{fig:dispersion_b_p75}
    \end{subfigure}
    \caption{Predicted and ground-truth dispersion bands for representative unseen binary geometries spanning the 25th, 50th, and 75th percentiles of prediction performance, restricted to the horizontal, vertical, and diagonal IBZ traversals.}
    \label{fig:dispersion_b_percentiles}
\end{figure}
As in the continuous geometry figures above, these plots show eigenfrequency predictions across all 325 wavevectors of the IBZ half-plane for binary geometries, restricted to the horizontal, vertical, and diagonal IBZ traversals. Overall, the model again achieves an average prediction error below $1\%$ on the encoded eigenfrequency across unseen samples.

\begin{figure}[H]
    \centering
    \begin{subfigure}[t]{0.49\textwidth}
        \centering
        \includegraphics[width=\textwidth]{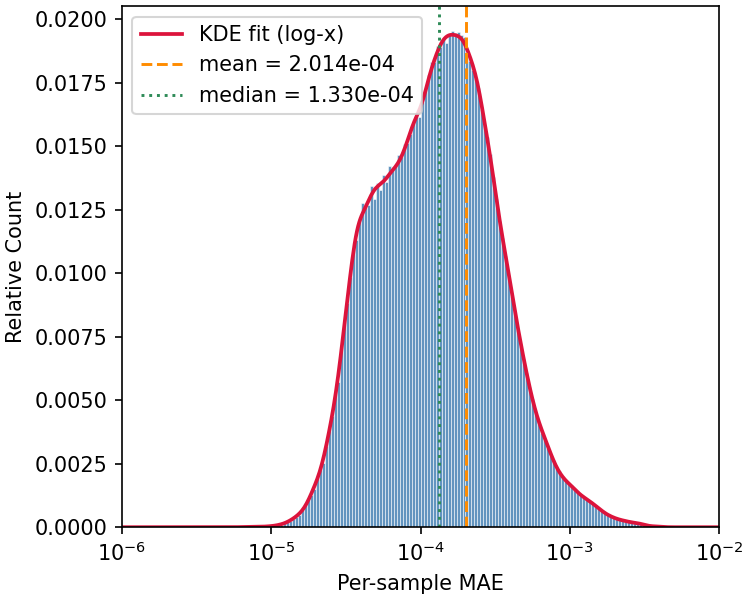}
        \caption{Performance histogram for continuous geometries.}
        \label{fig:freq_loss_histogram_continuous}
    \end{subfigure}
    \hfill
    \begin{subfigure}[t]{0.49\textwidth}
        \centering
        \includegraphics[width=\textwidth]{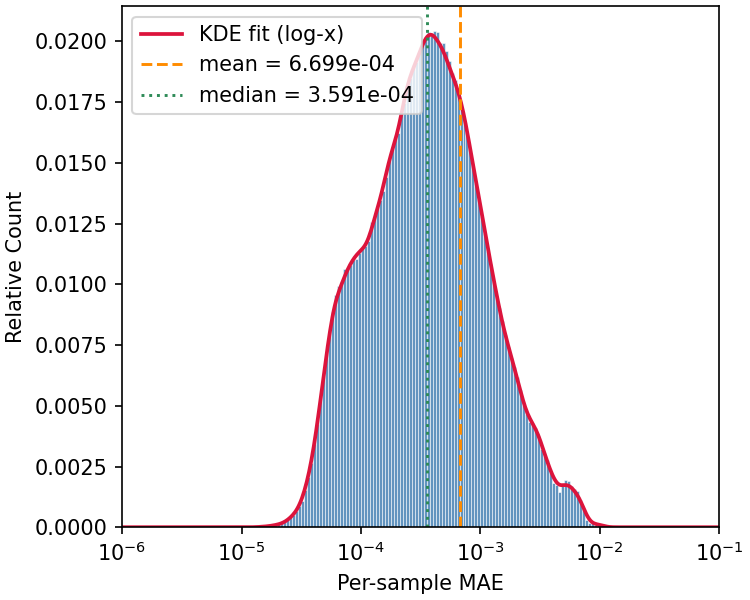}
        \caption{Performance histogram for binary geometries.}
        \label{fig:freq_loss_histogram_discrete}
    \end{subfigure}
    \caption{Comparison of relative prediction error distributions for continuous and binary metamaterial geometries for the same model. While there are some tail outliers, most samples fall within the range of $~10^{-2}$ to $~10^{-4}$, indicating that the same model is able to learn continuous and binary geometry cases.}
    \label{fig:freq_loss_histograms}
\end{figure}

\subsection{Dependence of Prediction Error on Band and Boundary Length}
\label{ssec:boundary_length_vs_loss}

\begin{figure}[H]
  \centering
  \includegraphics[width=\textwidth]{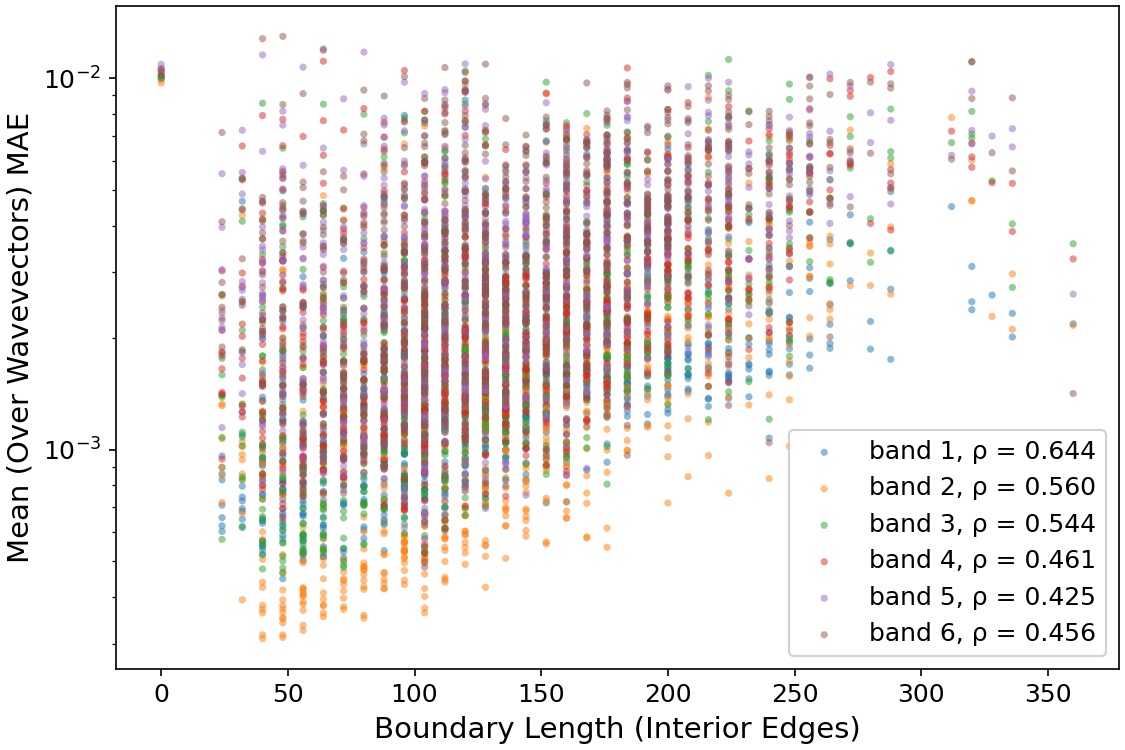}
  \caption{Mean displacement MAE versus interface boundary length, separated by color for each of the six eigenbands on the binary test set. Points correspond to average performance on unit-cell geometries. Legend values $\rho$ are Spearman rank correlations. A positive trend appears for every band, consistent with degraded FNO accuracy as the total length of $0$/$1$ material interfaces increases.}
  \label{fig:boundary_vs_mae_by_band}
\end{figure}

The preceding subsections showed that, although the same wavelet-conditioned FNO can predict displacement fields for both continuous and binary unit cells, binary geometries are systematically more difficult, with absolute and relative errors larger (Figures \ref{fig:disp_loss_histograms}, \ref{fig:freq_loss_histograms}), and the harder quartiles of the binary test set degrading more visibly than their continuous counterparts. This gap is expected from the architecture of the FNO itself and from the well-documented preference of spectral networks for low-frequency content~\cite{rahaman2019spectral,qin2024spectralfno}. Each Fourier layer retains only a truncated set of spectral modes after the discrete FFT. Smooth continuous material fields, and the comparatively smooth displacement modes they induce, concentrate their energy in a modest number of low-to-moderate wavenumbers and are therefore well represented by that truncated Fourier basis. Binary geometries, by contrast, contain jump discontinuities at material interfaces. Under an FFT, such discontinuities produce a broadband spectrum whose coefficients decay slowly with wavenumber, so a substantial fraction of the high-frequency content needed to resolve sharp phase boundaries lies outside the retained modes; related analyses of Fourier-based solvers for discontinuous coefficients likewise highlight Gibbs-type artifacts and degraded high-wavenumber recovery~\cite{cavallazzi2026whno}. The spectral branch therefore sees a degraded, band-limited version of the interface, yielding Gibbs-type ringing or blur and a harder operator-learning problem precisely when geometric discontinuity is more severe.
If this spectral-truncation explanation is correct, then even among binary geometries the prediction error should increase with the boundary interface length. To test this empirically, we define the \emph{boundary length} of each binarized unit cell as the number of interior four-connected pixel edges separating a $0$-valued pixel from a $1$-valued pixel. This scalar is a discrete measure of interface complexity. For each of the $1000$ unseen binary test geometries, we compute the mean absolute error (MAE) of the predicted displacement channels, averaged over all $325$ IBZ wavevectors, separately for each of the six retained eigenbands. The resulting relationships are shown in Figure~\ref{fig:boundary_vs_mae_by_band}.

Two complementary trends are apparent. First, for every band there is a positive monotonic association between boundary length and MAE, with Spearman correlations ranging from $\rho=0.644$ for band~$1$ to $\rho\approx0.42$--$0.56$ for the higher bands. Geometries with short interfaces are concentrated at lower absolute error, while geometries with long, meandering phase boundaries systematically occupy the upper portion of the cloud. This supports the interpretation that the hard cases for the learned operator are those for which the truncated Fourier representation must resolve more extensive sharp material transitions on the fixed $32\times32$ grid.
Second, absolute error itself increases with band index. Averaged over the test geometries, the mean wavevector-averaged MAE rises from approximately $1.6\times10^{-3}$ for band~$1$ to approximately $3.8\times10^{-3}$ for band~$6$. Higher bands correspond to more rapidly oscillating Bloch modes that concentrate energy nearer material interfaces and therefore inherit more of the geometric difficulty encoded by boundary length. Together with the continuous versus binary comparisons above, Figure~\ref{fig:boundary_vs_mae_by_band} indicates that prediction difficulty on binary designs is organized both by band (modal complexity) and by interface measure (geometric complexity), as expected when a spectral surrogate retains only a portion of the broadband Fourier content generated by discontinuities.

\subsection{Band and Wavevector Encoding}
\label{ssec:band_wavevector_encoding}
For the FNO to ingest the conditioning constants of the eigenvalue PDE in equation \ref{eqn:eig_PDE}, namely the wavevectors $k_x$, $k_y$ and the band index $b$ that selects which eigenmode to return, these scalars must be represented as matrices or tensors compatible with the existing channel wise input structure. Alternatives that inject scalars through specialized pathways would require heavy architectural alterations. Because FNOs are already expressive operator approximators when inputs are presented as fields~\cite{li2021fno,kovachki2023neural}, we keep that interface and compare three field encodings of the conditioning constants: constant valued fields, spatial sinusoids, and the Gabor wavelet encoding described in Section \ref{ssec:input_wavelet_encoding}.

The constant field representation is the naive default. On pixel coordinates $(x,y)\in\{0,\ldots,S-1\}$ with $S=32$, wavevector and band were encoded as
\[
I_{k_x}(x,y)=\frac{k_x}{\pi},\qquad I_{k_y}(x,y)=\frac{k_y}{\pi},\qquad k_x,k_y\in[-\pi,\pi],
\]
\[
I_b(x,y)=\frac{b}{10},\qquad b\in\{1,2,3,4,5,6\}.
\]
Each field is spatially uniform: the normalized scalars $\frac{k_x}{\pi},\frac{k_y}{\pi}\in[-1,1]$ and $\frac{b}{10}\in[0.1,0.6]$ are broadcast over the $S\times S$ grid. Because the FFT of a constant field is zero everywhere except at the DC coefficient, the frequency-domain branch of each Fourier layer receives little non-trivial information to propagate beyond the first layer, so learning must rely mainly on the shallow spatial-domain branch. In practice this encoding is fragile: under some training hyperparameters, particularly high learning rates, training collapses to near-constant outputs that ignore the conditioning inputs. With more stable settings the model can still learn, but it underperforms the wavelet encoding. The best uniform-encoding results we obtained under matched conditions are reported in Tables~\ref{tab:encoding_ablation_median} and~\ref{tab:encoding_ablation_mean} below.

A spatial sinusoidal encoding restores unique and nondegenerate representations in the spectral domain and allows the FNO to learn the problem. On pixel coordinates $(x,y)\in\{0,\ldots,S-1\}$, wavevector and band were encoded as
\[
I_k(x,y)=\sin\!\left(\frac{2}{S}\bigl(k_x x+k_y y\bigr)\right),\qquad k_x,k_y\in[-\pi,\pi].
\]
\[
I_b(x,y)=\tfrac{1}{2}\left[\cos\!\left(\frac{2\pi b x}{S}\right)+\cos\!\left(\frac{2\pi b y}{S}\right)\right],\qquad b\in\{1,2,3,4,5,6\}.
\]
The wavevector field is a directed plane wave: its phase gradient is parallel to $(k_x,k_y)$, so wavefronts propagate along the physical wavevector, and the spatial frequency scales with $|\mathbf{k}|$. The prefactor $2/S$ places about one cycle across the patch at the IBZ boundary $|\mathbf{k}|=\pi$, which keeps neighboring $(k_x,k_y)$ pairs visually and spectrally distinct on the $S\times S$ grid while avoiding severe aliasing. A sine rather than a cosine is used so that the $\Gamma$ point $(\mathbf{k}=\mathbf{0})$ maps to the zero field instead of a constant DC patch.
However, because a sinusoid is spectrally sparse, with energy concentrated at only a few wavenumbers, the Fourier branch remains weakly utilized, and predictions exhibit a persistent grainy, under-resolved structure relative to ground truth. Increasing model depth and width did not remove this graininess, indicating that the bottleneck was the encoding rather than model capacity.

The Gabor wavelet encoding resolves this by design. Following Gabor's classical construction~\cite{gabor1946theory}, wavelets are constructed to trade certainty between the spatial and spectral domains: neither representation is infinitely sharp, but both remain informationally rich. As a result, the conditioning inputs populate both branches of each Fourier layer with usable structure, both branches can train and contribute, and the model yields the accurate, well resolved predictions reported above while enabling deterministic mode selection. These observations support the broader conclusion that input encodings should be matched to the spatial and spectral structure of the operator: encodings that populate only one domain underuse the FNO's Fourier branch, whereas wavelet encodings exercise both.

To quantify these differences under matched training conditions, we compare wavelet-, sinusoidal-, and uniform-encoded models on held-out continuous and binary test geometries using overall-sample MAE, MSE, NMAE, and NMSE (Tables~\ref{tab:encoding_ablation_median} and~\ref{tab:encoding_ablation_mean}). All three runs use the same training hyperparameters: an FNO with 4 Fourier layers and 128 hidden channels, AdamW optimization, learning rate $2\times10^{-3}$ with StepLR step size 1 and per-epoch decay $0.9$, batch size 520, and an NMAE training objective. Checkpoints are taken after the 8th training epoch, where the sinusoidal- and uniform-encoded models begin to asymptote. The wavelet-encoded model continues to improve until roughly epoch 12, but the epoch-8 checkpoint is reported here for a fair comparison at equal training budget. Positive percentages in the tables indicate worse error than the wavelet-encoded model.

\begin{table}[H]
    \centering
    \small
    \begin{tabular}{llccccc}
        \toprule
        Geometry & Median & Wavelet & Sinusoidal & Versus & Uniform & Versus \\
        Dataset & Loss & Encoding & Encoding & Wavelet & Encoding & Wavelet \\
        \midrule
        \multirow{4}{*}{Continuous}
            & MAE  & $5.621\times10^{-4}$ & $7.393\times10^{-4}$ & $+31.6\%$ & $7.156\times10^{-4}$ & $+27.3\%$ \\
            & MSE  & $1.236\times10^{-6}$ & $1.825\times10^{-6}$ & $+47.7\%$ & $1.848\times10^{-6}$ & $+49.5\%$ \\
            & NMAE & $9.13\times10^{-2}$  & $1.202\times10^{-1}$ & $+31.6\%$ & $1.124\times10^{-1}$ & $+23.1\%$ \\
            & NMSE & $7.971\times10^{-3}$ & $1.135\times10^{-2}$ & $+42.4\%$ & $1.169\times10^{-2}$ & $+46.7\%$ \\
        \midrule
        \multirow{4}{*}{Binary}
            & MAE  & $1.336\times10^{-3}$ & $1.488\times10^{-3}$ & $+11.4\%$ & $1.520\times10^{-3}$ & $+13.8\%$ \\
            & MSE  & $5.004\times10^{-6}$ & $5.712\times10^{-6}$ & $+14.2\%$ & $6.150\times10^{-6}$ & $+22.9\%$ \\
            & NMAE & $1.800\times10^{-1}$ & $1.996\times10^{-1}$ & $+10.9\%$ & $2.051\times10^{-1}$ & $+13.9\%$ \\
            & NMSE & $3.01\times10^{-2}$  & $3.38\times10^{-2}$  & $+12.2\%$ & $3.76\times10^{-2}$  & $+24.7\%$ \\
        \bottomrule
    \end{tabular}
    \caption{Overall-sample median test losses and relative change versus the wavelet-encoded model. Lower is better.}
    \label{tab:encoding_ablation_median}
\end{table}

\begin{table}[H]
    \centering
    \small
    \begin{tabular}{llccccc}
        \toprule
        Geometry & Mean & Wavelet & Sinusoidal & Versus & Uniform & Versus \\
        Dataset & Loss & Encoding & Encoding & Wavelet & Encoding & Wavelet \\
        \midrule
        \multirow{4}{*}{Continuous}
            & MAE  & $1.405\times10^{-3}$ & $1.649\times10^{-3}$ & $+17.4\%$ & $1.591\times10^{-3}$ & $+13.2\%$ \\
            & MSE  & $3.234\times10^{-5}$ & $3.496\times10^{-5}$ & $+8.1\%$  & $3.529\times10^{-5}$ & $+9.1\%$ \\
            & NMAE & $2.876\times10^{-1}$ & $3.559\times10^{-1}$ & $+23.7\%$ & $3.466\times10^{-1}$ & $+20.5\%$ \\
            & NMSE & $1.405\times10^{-1}$ & $1.615\times10^{-1}$ & $+14.9\%$ & $1.619\times10^{-1}$ & $+15.3\%$ \\
        \midrule
        \multirow{4}{*}{Binary}
            & MAE  & $2.662\times10^{-3}$ & $2.830\times10^{-3}$ & $+6.3\%$  & $2.861\times10^{-3}$ & $+7.5\%$ \\
            & MSE  & $6.641\times10^{-5}$ & $6.756\times10^{-5}$ & $+1.7\%$  & $7.030\times10^{-5}$ & $+5.8\%$ \\
            & NMAE & $4.905\times10^{-1}$ & $5.416\times10^{-1}$ & $+10.4\%$ & $5.585\times10^{-1}$ & $+13.9\%$ \\
            & NMSE & $3.435\times10^{-1}$ & $3.500\times10^{-1}$ & $+1.9\%$  & $3.776\times10^{-1}$ & $+9.9\%$ \\
        \bottomrule
    \end{tabular}
    \caption{Overall-sample mean test losses and relative change versus the wavelet-encoded model. Lower is better.}
    \label{tab:encoding_ablation_mean}
\end{table}

\section{Conclusions}
\label{sec:Conclusion}
This work shows that a single Fourier Neural Operator can learn multiple eigenmodes of the elastic wave equation for acoustic metamaterials, recovering Bloch displacement fields and eigenfrequencies across both continuous and binary unit cell geometries. Relative to prior surrogates that mainly target dispersion curves, transmission spectra, or bandgap scalars~\cite{Ogren2024gpr,wagner2024neural,liu2026hybridfno}, the present model returns full deformation modes with deterministic selection among modes through wavelet encodings of the wavevector and band index. On unseen geometries of both types, displacement predictions typically achieve relative errors between $10^{-2}$ and $10^{-4}$, and reconstructed dispersion relations average below $1\%$ error. Encoding choice is essential for this capability. Constant field encodings fail to propagate information through spectral branches of the FNO, leading to poor model performance. Sinusoidal encodings train better but remain grainy due to unbalanced information allocation between spatial and spectral branches. Gabor wavelet encodings~\cite{gabor1946theory} supply rich information structure in both the spatial and spectral branches of the FNO, enabling more reliable mode selection and learning.

A fundamental limitation of FNOs is its prediction error tends to grow with discontinuities in its inputs and outputs, as evidenced by decreasing performance with increasing interface length. Despite this, performance remains high fidelity for the trained surrogate, which is lightweight (about $2\,\mathrm{GB}$ at float16) and evaluates in about $1\,\mathrm{ms}$ per sample on a consumer-grade CPU, compared with about $1\,\mathrm{s}$ per sample for the corresponding FEA eigenvalue solve. That three orders of magnitude speedup makes repeated Bloch analyses, which are central to optimization, uncertainty quantification, inverse design, and screening of unit cell geometries, far more practical than direct finite element evaluation during design iterations. In short, FNOs with wavelet encodings of modal parameters provide a pathway to greatly accelerate acoustic metamaterial simulation and design relative to repeated finite element eigenvalue analysis. Future work includes other spatial resolutions, physics informed operator losses~\cite{wang2021pidon}, and broader design spaces beyond the present eight fold symmetric, two material unit cells, including other tiling symmetries, asymmetric geometries, and multimaterial compositions.

\section*{AI Usage Disclosure}
Generative AI tools were used to assist with coding of experiments, translating MATLAB code to Python, and automating training pipelines. AI tools were also used for light editing of the manuscript. All scientific claims, results, and final wording were reviewed and approved by the authors.

\section*{Declaration of Competing Interest}
The authors declare that they have no known competing financial interests or personal relationships that could have appeared to influence the work reported in this paper.

\section*{CRediT authorship contribution statement}
\noindent\textbf{Han Zhang:} Conceptualization, Data curation, Formal analysis, Investigation, Methodology, Software, Validation, Visualization, Writing -- original draft, Writing -- review and editing.\\
\noindent\textbf{Alexander Ogren:} Data curation, Software.\\
\noindent\textbf{Cynthia Rudin:} Supervision, Writing -- review and editing.\\
\noindent\textbf{Johann Guilleminot:} Formal analysis, Methodology, Supervision, Writing -- review and editing.\\
\noindent\textbf{L. Catherine Brinson:} Conceptualization, Funding acquisition, Methodology, Project administration, Resources, Software, Supervision, Visualization, Writing -- review and editing.

\section*{Data availability}
\label{sec:Repository}
Python code and experiments can be found at
\url{https://github.com/trutheresy/NO-2D-Metamaterials}.

MATLAB FEA code can be found at
\url{https://github.com/aco8ogren/2D-dispersion.git}.

\section*{Acknowledgements}
This work was supported by the NSERC Postgraduate Scholarship -- Doctoral (PGSD-599307-2025) and the NSF AI for Understanding and Designing Materials Research Traineeship (DGE-2022040).

\bibliographystyle{elsarticle-num}
\bibliography{references}

\clearpage
\appendix

\renewcommand{\thesection}{S\arabic{section}}
\renewcommand{\thesubsection}{S\arabic{section}.\arabic{subsection}}
\renewcommand{\thefigure}{S\arabic{figure}}
\renewcommand{\thetable}{S\arabic{table}}
\renewcommand{\theequation}{S\arabic{equation}}

\setcounter{figure}{0}
\setcounter{table}{0}
\setcounter{equation}{0}

\section{Supplementary Information}
\label{sec:supplement}
\subsection{Model and Training Hyperparameters}
\begin{table}[H]
    \centering
    \begin{tabular}{|c|c|c|c|c|}
        \hline
        \textbf{Hyperparameter} &
        \multicolumn{4}{c|}{\textbf{Candidate Settings}} \\ \hline
    
        \textbf{Fourier Layers} & \textit{4} & 6 & 8 & 12 \\ \hline
        \textbf{Hidden Channels} & 32 & 64 & \textit{128} & 256 \\ \hline
        \textbf{Activation Function} & ReLU & \textit{GELU} & tanh & Sigmoid \\ \hline
        \textbf{Loss Criterion} & L1 & L2 & \textit{NMAE} & NMSE \\ \hline
        \textbf{Epochs} & 8 & \textit{12} & 16 & 20 \\ \hline
        \textbf{Learning Rate} & $10^{-4}$ & \textit{$2\times10^{-3}$} & $10^{-2}$ & $10^{-1}$ \\ \hline
        \textbf{Weight Decay} & \textit{0} & $10^{-4}$ & $10^{-3}$ & $10^{-2}$ \\ \hline
        \textbf{Gamma} & 0 & 0.01 & 0.1 & \textit{0.9} \\ \hline
        \textbf{Step Size} & 0 & \textit{1} & 4 & 10 \\ \hline
        \textbf{Batch Size} & 64 & 128 & 256 & \textit{520} \\ \hline
        \textbf{Data Precision} & F8 E4M3 & F8 E5M2 & \textit{F16} & F32 \\ \hline
    \end{tabular}
    \caption{Combinations tried in grid search for optimal model and training hyperparameters. Each row is a set of hyperparameters that can be paired with any from other rows.}
    \label{tab:training_hyperparameters}
\end{table}

\subsection{Loss Criterion}
In the exploration of optimal training protocols for the FNO model, several loss functions were evaluated as training objectives: mean absolute error (MAE), mean squared error (MSE), normalized MAE (NMAE), normalized MSE (NMSE), and structural similarity index (SSIM). Among these, NMAE yielded the best overall performance and was used for all reported results. All comparisons used a learning rate of $2\times10^{-3}$ with a per-epoch decay factor of $0.9$. The mathematical definitions of each loss follow. Losses were evaluated and aggregated across all five output channels. 

\begin{itemize}
    \item The mean absolute error (MAE) loss is defined as
\begin{equation*}
    \mathcal{L}_{\text{MAE}} = \frac{1}{HW} \sum_{i=0}^{H-1} \sum_{j=0}^{W-1} \sum_{c=0}^{4} \left| \hat{u}_{c}(i,j) - u_{c}(i,j) \right|
\end{equation*}

Experimentally, models trained exclusively with MAE produced predictions with sharper interfaces and better-preserved boundary features than those trained solely with MSE. This behavior is consistent with the linear penalty of MAE, which does not disproportionately penalize isolated large residuals near discontinuities. Despite the improved qualitative sharpness, MAE training underperformed NMAE on overall validation accuracy, so absolute-error training alone was not selected for the reported models.

\item The mean squared error (MSE) loss is defined as
\begin{equation*}
    \mathcal{L}_{\text{MSE}} = \frac{1}{HW} \sum_{i=0}^{H-1} \sum_{j=0}^{W-1} \sum_{c=0}^{4} \left( \hat{u}_{c}(i,j) - u_{c}(i,j) \right)^2
\end{equation*}

The quadratic penalty of MSE assigns increasingly large gradients to larger residuals, causing the optimizer to preferentially reduce regions with the greatest numerical error. As a consequence, the loss favors smooth predictions that minimize the overall energy of the residual rather than preserving sharp transitions. Experimentally, models trained exclusively with MSE produced outputs that reproduced the correct magnitude and large-scale structure of the solution, but exhibited diffuse interfaces and blurred boundaries. Fine-scale features were often replaced by low-amplitude, grainy artifacts, consistent with the tendency of MSE to average over uncertain or discontinuous regions in order to minimize the global squared error.

\item The structural similarity index measure (SSIM) is defined as

\begin{equation*}
\mathcal{L}_{\text{SSIM}}
=
1
-
\frac{1}{5}
\sum_{c=0}^{4}
\frac{
\left(2\mu_{u_c}\mu_{\hat{u}_c}+C_1\right)\left(2\sigma_{u_c\hat{u}_c}+C_2\right)
}{
\left(\mu_{u_c}^2+\mu_{\hat{u}_c}^2+C_1\right)\left(\sigma_{u_c}^2+\sigma_{\hat{u}_c}^2+C_2\right)
}, \quad \text{where}
\end{equation*}
\begin{equation*}
\mu_{u_c}=\frac{1}{HW}\sum_{i=0}^{H-1}\sum_{j=0}^{W-1}u_c(i,j), \quad
\sigma_{u_c\hat{u}_c}=\frac{1}{HW}\sum_{i=0}^{H-1}\sum_{j=0}^{W-1}\left(u_c(i,j)-\mu_{u_c}\right)\left(\hat{u}_c(i,j)-\mu_{\hat{u}_c}\right)
\end{equation*}

A significant limitation of SSIM arises when it is applied to signed-valued fields. Unlike its original application to nonnegative image intensities, a perfect sign-reversed field ($\hat{u}=-u$) can achieve nearly the same SSIM score as a perfect reconstruction ($\hat{u}=u$). Consequently, a model trained solely with SSIM may converge to physically incorrect solutions containing polarity inversions while still minimizing the loss. The following derivation demonstrates the mathematical origin of this ambiguity.

For the purposes of the following discussion, we consider the common case where the stabilizing constants satisfy
\begin{equation*}
    C_1 \ll 2\mu_{u_c}^2,
    \qquad
    C_2 \ll 2\sigma_{u_c}^2,
\end{equation*}
allowing them to be neglected.

For a correct reconstruction,
\begin{equation*}
    u_c^{(+)} = u_c,
\end{equation*}
the SSIM score is
\begin{equation*}
\mathrm{SSIM}(u_c,u_c^{(+)})
=
\frac{\left(2\mu_{u_c}^2\right)\left(2\sigma_{u_c}^2\right)}
{\left(2\mu_{u_c}^2\right)\left(2\sigma_{u_c}^2\right)}
=1.
\end{equation*}

Now consider a sign-reversed reconstruction. The field and its mean reverse sign,
\begin{equation*}
    u_c^{(-)}=-u_c, \qquad \mu_{u_c^{(-)}}=-\mu_{u_c},
\end{equation*}
while the covariance reverses sign and the variance remains unchanged,
\begin{equation*}
    \sigma_{u_c,u_c^{(-)}}=-\sigma_{u_c,u_c}, \qquad \sigma_{u_c^{(-)}}^2=\sigma_{u_c}^2.
\end{equation*}
Substituting these relationships into the SSIM expression gives
\begin{equation*}
\mathrm{SSIM}(u_c,u_c^{(-)})
=
\frac{\left(-2\mu_{u_c}^2\right)\left(-2\sigma_{u_c}^2\right)}
{\left(2\mu_{u_c}^2\right)\left(2\sigma_{u_c}^2\right)}=1.
\end{equation*}
The numerator therefore differs from that of a correct reconstruction by two factors of $-1$, and thus, under these conditions, SSIM assigns essentially the same similarity score to a field and its sign-reversed counterpart as it does to a perfect reconstruction. This ambiguity constitutes a significant limitation when SSIM is applied to signed-valued regression problems where the polarity of the solution is physically meaningful. During training, the network frequently converged to predictions in which localized regions or the entire image were the negative of the correct solution while still achieving a favorable SSIM objective. Although the resulting fields exhibited accurate interface locations, boundary geometry, and overall morphology, the polarity inversions rendered the predictions physically incorrect. Consequently, SSIM should not be used as the sole training criterion for signed-valued fields unless it is supplemented by an additional loss that explicitly penalizes sign reversals.

\item The normalized mean absolute error (NMAE) loss is defined as
\begin{equation*}
    \mathcal{L}_{\text{NMAE}} = \frac{1}{HW} \sum_{i=0}^{H-1} \sum_{j=0}^{W-1} \sum_{c=0}^{4} \frac{\displaystyle \left| \hat{u}_{c}(i,j) - u_{c}(i,j) \right|}{\left| u_{c}(i,j) \right| + \varepsilon}
\end{equation*}
Unlike MAE, NMAE weights each prediction error by the inverse magnitude of the target value, causing the optimization to place greater emphasis on regions where the ground-truth solution is small in scale. Consequently, the loss seeks to minimize relative error rather than absolute error, preventing low-magnitude features from being dominated by high-magnitude regions during training.

Experimentally, NMAE gave the best overall balance across channels with disparate magnitudes and was selected as the training objective for all reported results. Relative-error emphasis improved reconstruction of low-amplitude displacement components without sacrificing usable accuracy on larger-magnitude channels, which made NMAE preferable to MAE, MSE, NMSE, and SSIM for this application.

\item The normalized mean squared error (NMSE) loss is defined as
\begin{equation*}
    \mathcal{L}_{\text{NMSE}} = \frac{1}{HW} \sum_{i=0}^{H-1} \sum_{j=0}^{W-1} \sum_{c=0}^{4} \frac{\displaystyle \left(\hat{u}_{c}(i,j) - u_{c}(i,j) \right)^2}{\left( u_{c}(i,j) \right)^2 + \varepsilon}
\end{equation*}

Similar to NMAE, NMSE normalizes the prediction error by the magnitude of the target field, causing the optimization to prioritize relative error rather than absolute error. The quadratic penalty retains the characteristic behavior of MSE by assigning greater weight to larger residuals, while the normalization emphasizes regions where the target magnitude is small.

Experimentally, NMSE reduced the relative error in low-amplitude regions compared with standard MSE, but it underperformed NMAE on overall validation accuracy and was not selected for the reported models.
\end{itemize}

\subsection{Algorithms for Input Wavelet Encoding}
\label{ssec:input_wavelet_algorithms}
The band and wavevector conditioning channels used by the FNO are generated by the Gabor wavelet embeddings below, which match the implementations in the accompanying code.
\begin{algorithm}[H]
\caption{1D Gabor Wavelet Embedding from Band Index}
\label{alg:1d_gabor_embedding}
\begin{algorithmic}[1]
\STATE \textbf{INPUT:} Band index \( b \); grid size \( S \leftarrow 32 \); frequency scale \( r \leftarrow 2.0 \)
\STATE Create linspace vectors \( x, y \in [-1, 1] \) with \( S \) points
\STATE Construct meshgrid \( X, Y \leftarrow \text{meshgrid}(x, y) \)
\STATE Compute base frequency:
\[
f \leftarrow (1.0 + |b|) \cdot \frac{S}{8}
\]
\STATE Compute rotation angle:
\[
\theta \leftarrow (b \bmod 8) \cdot \frac{S}{16}
\]
\STATE Rotate coordinates:
\[
X_\theta \leftarrow X \cos \theta + Y \sin \theta, \quad
Y_\theta \leftarrow -X \sin \theta + Y \cos \theta
\]
\STATE Set Gaussian envelope widths:
\[
\sigma_x \leftarrow \sigma_y \leftarrow \frac{0.3}{r}
\]
\STATE Compute Gabor wavelet embedding:
\[
\psi_b(x, y) \leftarrow
\exp\left(-\frac{X_\theta^2}{2 \sigma_x^2} - \frac{Y_\theta^2}{2 \sigma_y^2}\right)
\cdot \cos(f \cdot X_\theta)
\]
\STATE \textbf{RETURN:} \( \psi_b \in \mathbb{R}^{S \times S} \)
\end{algorithmic}
\end{algorithm}

In Algorithm \ref{alg:1d_gabor_embedding}, the grid size \(S=32\) matches the geometry resolution used by the FNO. For this resolution the base frequency is \(f=(1+|b|)\,S/8\), so successive bands increase both carrier frequency and orientation. The modulo in \(\theta\) repeats orientation every eight bands, while \(f\) continues to increase, so embeddings remain distinguishable beyond eight bands. The constants \(r=2.0\), \(S/16\), and \(0.3\) were chosen so that the wavelets occupy most of the spatial domain and remain visually distinct across bands.

\begin{algorithm}[H]
\caption{2D Gabor Wavelet Embedding for Wavevectors}
\label{alg:2d_gabor_embedding}
\begin{algorithmic}[1]
\STATE \textbf{INPUT:} Wavevector components \( k_x, k_y \) (radians); grid size \( S \leftarrow 32 \); frequency scale \( r \leftarrow 1.0 \)
\STATE Set constants \( f_0 \leftarrow 2.20 \), \( \alpha \leftarrow 41 \), \( N_x \leftarrow 13 \), \( N_y \leftarrow 25 \)
\STATE Create linspace vectors \( x, y \in [-1, 1] \) with \( S \) points
\STATE Construct meshgrid \( X, Y \leftarrow \text{meshgrid}(x, y) \)
\STATE Compute frequencies:
\[
f_x \leftarrow (f_0 + k_x)\,\alpha, \quad f_y \leftarrow (f_0 + k_y)\,\alpha
\]
\STATE Compute rotation angles:
\[
\theta_x \leftarrow (k_x \bmod N_x) \cdot \frac{\pi}{N_x}, \quad \theta_y \leftarrow (k_y \bmod N_y) \cdot \frac{\pi}{N_y}
\]
\STATE Rotate coordinates:
\[
X_\theta \leftarrow X \cos \theta_x + Y \sin \theta_y, \quad
Y_\theta \leftarrow -X \sin \theta_x + Y \cos \theta_y
\]
\STATE Set Gaussian envelope widths:
\[
\sigma_x \leftarrow \sigma_y \leftarrow \frac{0.5}{r}
\]
\STATE Compute Gabor wavelet embedding:
\[
\psi_{k_x, k_y}(x, y) \leftarrow
\exp\left(-\frac{X_\theta^2}{2 \sigma_x^2} - \frac{Y_\theta^2}{2 \sigma_y^2}\right)
\cdot \sin(f_x X_\theta) \cdot \sin(f_y Y_\theta)
\]
\STATE \textbf{RETURN:} \( \psi_{k_x, k_y} \in \mathbb{R}^{S \times S} \)
\end{algorithmic}
\end{algorithm}

In Algorithm \ref{alg:2d_gabor_embedding}, \(k_x\) and \(k_y\) enter as continuous radian valued wavevector components. Distinct modulo periods \(N_x=13\) and \(N_y=25\) reduce aliasing under interchange of the two components. The offset \(f_0\) and scale \(\alpha\) map the physical range of \(k\) into carrier frequencies that remain well resolved on the \(S\times S\) grid, while \(\sigma=0.5/r\) sets the envelope width so that the patterns remain spatially localized yet informative in both domains.

\subsection{Wavelet Decoding Fidelity}
\label{ssec:wavelet_decoding}
\begin{figure}[H]
  \centering
  \includegraphics[width=\textwidth]{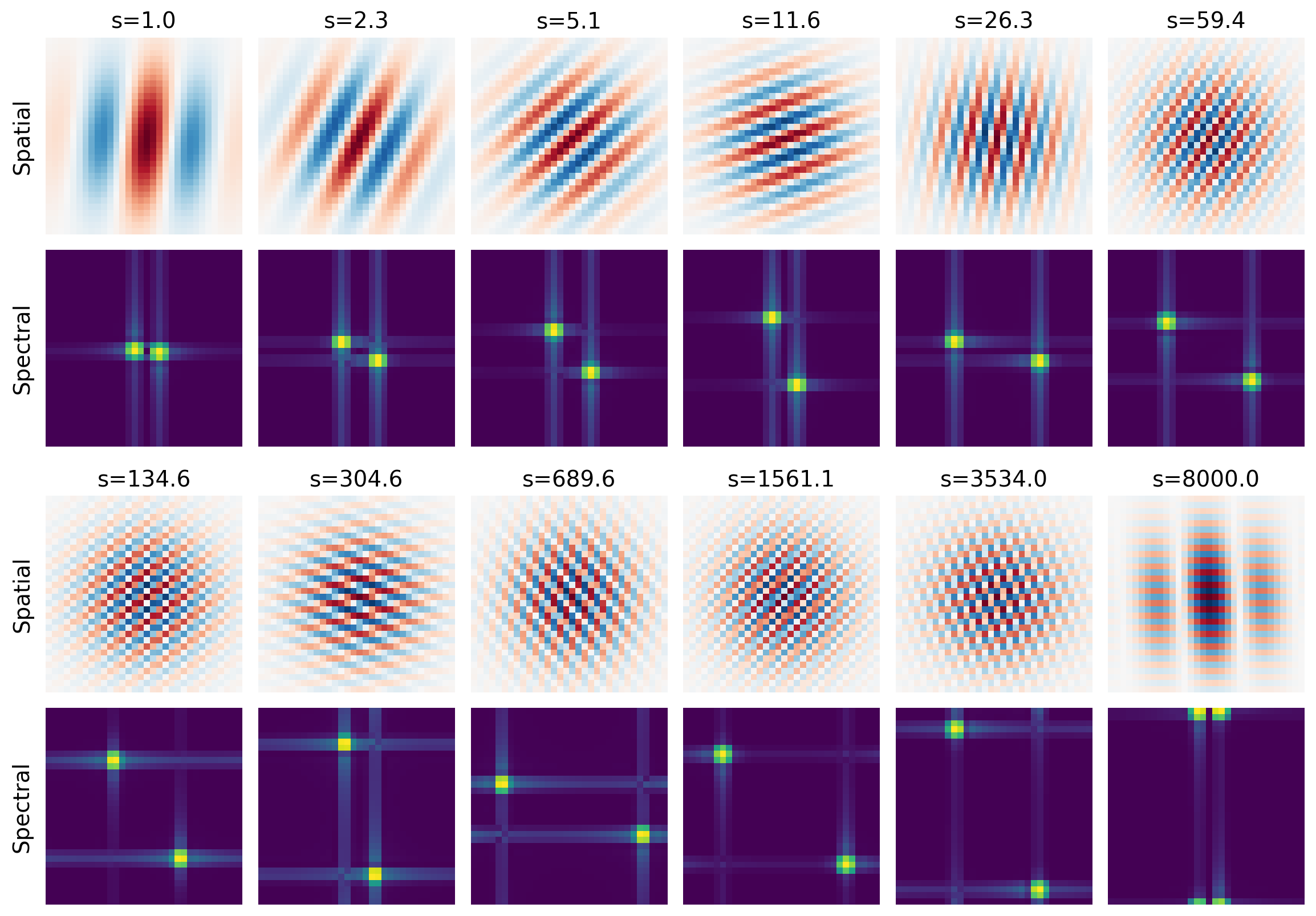}
  \caption{Log spaced Gabor wavelet encodings of scalars from 1 to 8000. Alternating rows show the spatial wavelet images and the corresponding Fourier magnitude spectra.}
  \label{fig:eigenfrequency_encoding}
\end{figure}

\begin{figure}[H]
  \centering
  \includegraphics[width=\textwidth]{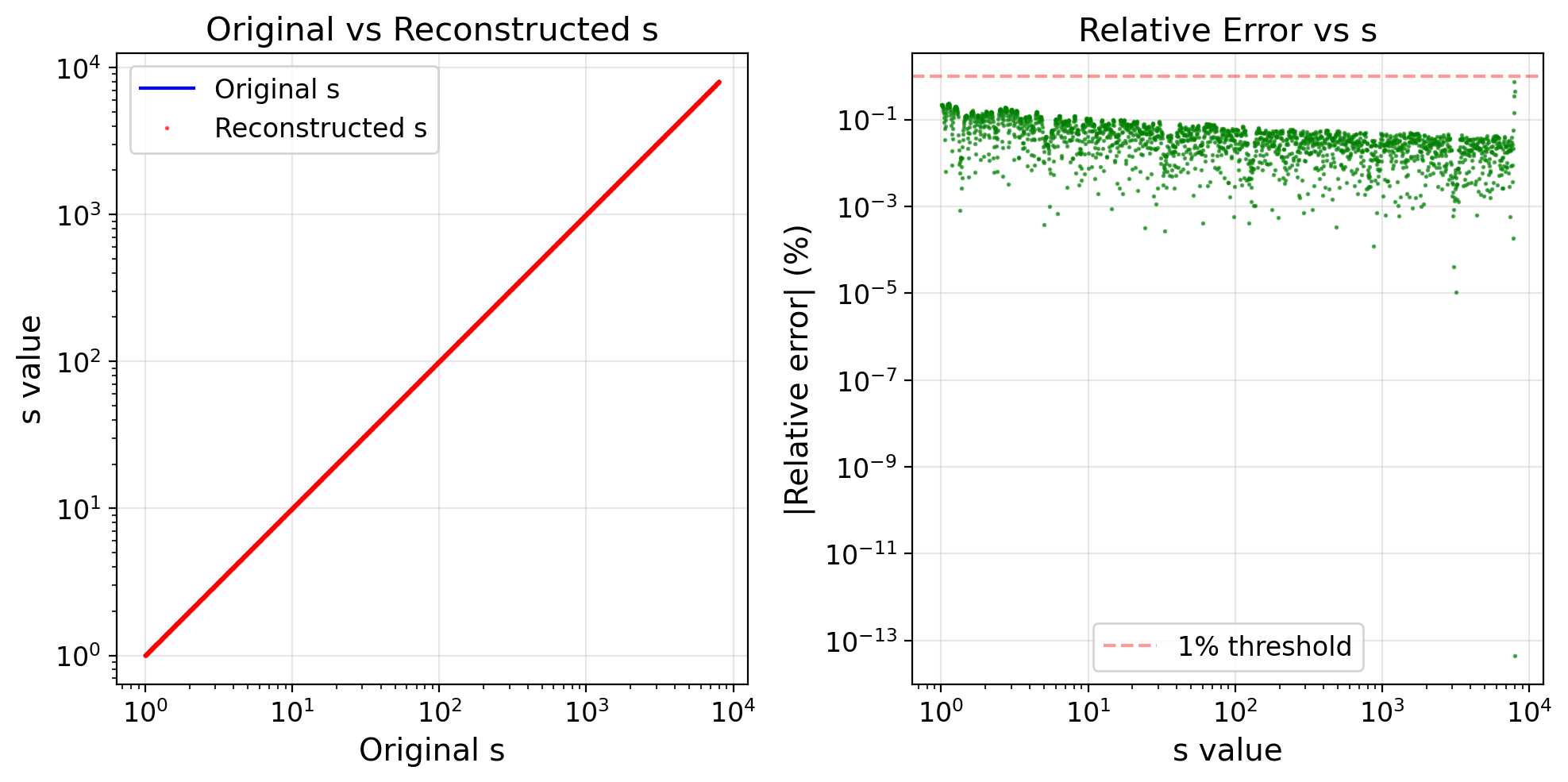}
  \caption{Encode--decode error for scalars from a minimum of 1 to a maximum of 8000.}
  \label{fig:eigenfrequency_decode_error}
\end{figure}

For purposes of checking general applicability, an experiment was run to check if a positive scalar may be embedded in a Gabor wavelet image and recovered after pixelization, storage, or inference. Algorithms \ref{alg:eigenfrequency_wavelet_encoding} and \ref{alg:eigenfrequency_wavelet_decoding} implement this encode--decode pair by spectral peak detection with centroid refinement. In the experiment, we test recovery over \([1,8000]\), spanning about four orders of magnitude. Despite discretization onto a finite \(S\times S\) grid and limited numeric precision associated with float16 storage and arithmetic, the decoded values remain accurate across the full range. Figures \ref{fig:eigenfrequency_encoding} and \ref{fig:eigenfrequency_decode_error} show representative encodings and the corresponding round trip error.

Algorithms \ref{alg:eigenfrequency_wavelet_encoding} and \ref{alg:eigenfrequency_wavelet_decoding} summarize the forward and inverse mapping used in this experiment. The forward mapping places the scalar on a log scale and jointly encodes magnitude and orientation into a Gabor-like pattern. The inverse mapping recovers an estimate by spectral peak detection with centroid refinement, followed by a consistency step that combines extracted magnitude and angle to reconstruct the most likely log value.

\begin{algorithm}[H]
\caption{Scalar Wavelet Encoding}
\label{alg:eigenfrequency_wavelet_encoding}
\begin{algorithmic}[1]
\STATE \textbf{INPUT:} Scalar \( s>0 \); image size \( S \leftarrow 32 \); bounds \( [s_{\min}, s_{\max}] \); spectral-index bounds \( [k_{\min}, k_{\max}] \)
\STATE Set constants \( \gamma \leftarrow 1,\ \phi \leftarrow 0,\ \sigma \leftarrow 8,\ \theta_{\min}\leftarrow \pi/30,\ \theta_{\max}\leftarrow \pi/2-\pi/30 \)
\STATE Compute log-domain values:
\[
\ell_s \leftarrow \log s,\quad \ell_{\min}\leftarrow \log s_{\min},\quad \ell_{\max}\leftarrow \log s_{\max}
\]
\STATE Compute normalized position and spectral index:
\[
t \leftarrow \mathrm{clip}\!\left(\frac{\ell_s-\ell_{\min}}{\ell_{\max}-\ell_{\min}},0,1\right),\quad
k \leftarrow k_{\min} + t\,(k_{\max}-k_{\min})
\]
\STATE Compute log width per unit \(k\):
\[
\Delta_\ell \leftarrow \frac{\ell_{\max}-\ell_{\min}}{k_{\max}-k_{\min}}
\]
\STATE Compute orientation from within-band log position:
\[
\theta \leftarrow \theta_{\min} + \left(\frac{(\ell_s-\ell_{\min})\bmod \Delta_\ell}{\Delta_\ell}\right)\left(\theta_{\max}-\theta_{\min}\right)
\]
\STATE Build centered grid \(X,Y \in \{-S/2,\dots,S/2-1\}\), then rotate:
\[
X_\theta \leftarrow X\cos\theta + Y\sin\theta,\quad
Y_\theta \leftarrow -X\sin\theta + Y\cos\theta
\]
\STATE Set envelope widths \( \sigma_x \leftarrow \sigma,\ \sigma_y \leftarrow \gamma \sigma_x \), and carrier frequency \( \omega \leftarrow 2\pi k/S \)
\STATE Compute wavelet image:
\[
\psi_s \leftarrow
\exp\!\left(-\frac{1}{2}\left(\frac{X_\theta^2}{\sigma_x^2}+\frac{Y_\theta^2}{\sigma_y^2}\right)\right)\cos(\omega X_\theta+\phi)
\]
\STATE \textbf{RETURN:} \( \psi_s \in \mathbb{R}^{S\times S},\ k,\ \theta \)
\end{algorithmic}
\end{algorithm}

\begin{algorithm}[H]
\caption{Scalar Wavelet Decoding}
\label{alg:eigenfrequency_wavelet_decoding}
\begin{algorithmic}[1]
\STATE \textbf{INPUT:} Wavelet image \( \psi \in \mathbb{R}^{S\times S} \); bounds \( [s_{\min}, s_{\max}] \), \( [k_{\min}, k_{\max}] \), \( [\theta_{\min}, \theta_{\max}] \)
\STATE Set constants for centroid refinement: radius \(R\leftarrow 3\), power \(p\leftarrow 2\)
\STATE Mean-center image \( \psi_c \leftarrow \psi - \mathrm{mean}(\psi) \), then compute the centered 2D discrete Fourier magnitude:
\[
\hat{\psi}(k_x,k_y)\leftarrow \sum_{m=0}^{S-1}\sum_{n=0}^{S-1}\psi_c(m,n)\,
\exp\!\left(-2\pi i\!\left(\frac{k_x m}{S}+\frac{k_y n}{S}\right)\right),\quad
F(k_x,k_y)\leftarrow |\hat{\psi}(k_x,k_y)|
\]
\STATE Keep upper-half-plane spectral support (plus positive-frequency center row) and find dominant peak index \( (r^{*},c^{*}) \)
\STATE Define symmetric peak partner about center, then compute weighted local centroid around \( (r^{*},c^{*}) \) within radius \(R\), using weights \(w=F^p\) and excluding points closer to the symmetric partner
\STATE Convert refined centroid to wavevector coordinates \( (k_x,k_y) \), then:
\[
k_{\mathrm{ext}} \leftarrow \sqrt{k_x^2+k_y^2},\quad
\theta_{\mathrm{ext}} \leftarrow \operatorname{atan2}(k_y,k_x)\bmod \pi
\]
\STATE Map extracted \(k\) to approximate log value:
\[
\ell_{\mathrm{approx}} \leftarrow \ell_{\min} + \mathrm{clip}\!\left(\frac{k_{\mathrm{ext}}-k_{\min}}{k_{\max}-k_{\min}},0,1\right)\,(\ell_{\max}-\ell_{\min})
\]
\STATE Let \( \Delta_\ell \leftarrow (\ell_{\max}-\ell_{\min})/(k_{\max}-k_{\min}) \), compute within-band position from angle:
\[
\alpha \leftarrow \mathrm{clip}\!\left(\frac{\theta_{\mathrm{ext}}-\theta_{\min}}{\theta_{\max}-\theta_{\min}},0,1\right),\quad
\delta_\ell \leftarrow \alpha\,\Delta_\ell
\]
\STATE Find nearest consistent log value by testing neighboring bands \(b-1,b,b+1\):
\[
\ell^\star \leftarrow \arg\min_{\ell \in \{\ell_{\min} + j\Delta_\ell + \delta_\ell\}} |\ell-\ell_{\mathrm{approx}}|
\]
\STATE Clip \( \ell^\star \) to \( [\ell_{\min},\ell_{\max}] \), then decode:
\[
s_{\mathrm{ext}} \leftarrow \exp(\ell^\star)
\]
\STATE \textbf{RETURN:} \( s_{\mathrm{ext}},\ k_{\mathrm{ext}},\ \theta_{\mathrm{ext}} \)
\end{algorithmic}
\end{algorithm}

\end{document}